%% file: main.tex
\documentclass[10pt,journal,compsoc]{IEEEtran}
\usepackage{blindtext}
\usepackage{graphicx}
\usepackage{tikz}

\usepackage{xcolor}

\usepackage{times}
\usepackage{epsfig}
\usepackage{graphicx}
\usepackage{amsmath}
\usepackage{amssymb}
\usepackage[breaklinks=true,bookmarks=false]{hyperref}
\usepackage{subcaption}
\usepackage{pdfpages}
\usepackage[linesnumbered,ruled]{algorithm2e}
\usepackage{todonotes}
\usepackage{algpseudocode}
\usepackage{epstopdf}

\newenvironment{edit}{\par}{\par}
\newcommand\mynotes[1]{#1}%{\textcolor{red}{#1}}

\usepackage{lipsum}
\makeatletter  % Edit the following values as appropriate
\def\@IEEEBIOphotowidth{0cm}    % width of the biography photo area 
\def\@IEEEBIOphotodepth{1cm}   % depth (height) of the biography photo area
\def\@IEEEBIOhangwidth{0cm}    % width cleared for the biography photo area
\def\@IEEEBIOhangdepth{0cm}    % depth cleared for the biography photo area 
\makeatother
\begin{document}
\title{Scattering Networks for Hybrid Representation Learning}

\author{Edouard Oyallon,
        Sergey Zagoruyko,
        Gabriel Huang,
        Nikos Komodakis,
        Simon Lacoste-Julien,
        Matthew Blaschko,
                Eugene Belilovsky

\thanks{\textit{Edouard Oyallon is with CentraleSupelec CVN and INRIA Galen team, France}}% <-this % stops a space
\thanks{\textit{Matthew Blaschko is affiliated with KU Leuven in Leuven, Belgium}}% <-this % stops a space
\thanks{\textit{Sergey Zagoruyko and Nikos Komodakis are at Ecole des Points in France}}
\thanks{\textit{Simon Lacoste-Julien, Eugene Belilovsky, and Gabriel Huang are members of MILA and DIRO at University of Montreal, Montreal, Canada}}
}

%\usepackage{todonotes}
%\usepackage{url}
%\usepackage{tikz}
%\usetikzlibrary{shapes.geometric}
\newtheorem{prop}{Proposition}

\newcommand{\citep}{\cite}

\newcommand{\addition}{} %\color{red}}

\markboth{IEEE TRANSACTIONS ON PATTERN ANALYSIS AND MACHINE INTELLIGENCE}%
{Shell \MakeLowercase{\textit{et al.}}: Bare Demo of IEEEtran.cls for Journals}



\maketitle

\input{intro_abstract}

%\section{Scattering Networks and Hybrid Architectures}
\section{Scattering: a  baseline for image classification}
\label{hyb}

\input{scattering_network}

\input{supervision_on_top}

%%%%%%%
\section{Local Encoding of Scattering}
\label{sle_sec}
First, we motivate the use of the Shared Local Encoder for natural image classifications. Then, we evaluate the supervised SLE on the Imagenet ILSVRC2012 dataset. This is a large and challenging natural color image dataset consisting of $1.2$ million training images and $50,000$ validation images, divided into $1000$ classes. We then show some unique properties of this network and evaluate its features on a separate task.

\subsection{Shared Local Encoder for Scattering Representations}
\label{supervised_encoding}
\input{supervised_encoding}

\subsection{Shared Local Encoder on Imagenet}
\label{sec:sle_exp}
\input{SLE.tex}

\subsection{Interpreting SLE's first layer}
\label{sec:sle_first_sec}
 %of the Scatter Encodings}\label{simpl}}
\label{interpretation}

\input{interpretating.tex}
\section{Cascading a Supervised Deep CNN Architecture}
We  demonstrate that cascading modern CNN architectures on top of the scattering network can produce high performance classification systems. We apply hybrid convolutional networks on the Imagenet ILSVRC 2012 dataset as well as the CIFAR-10 dataset and show that they can achieve performance comparable to modern end-to-end learned approaches. We then evaluate the hybrid networks in the setting of limited data by utilizing a subset of CIFAR-10 as well as the STL-10 dataset and show that we can obtain substantial improvement in performance over analogous end-to-end learned CNNs.

\label{small2}
\input{small}

%%%%%%%
\addition
\section{Unsupervised and Hybrid Unsupervised Learning with the Scattering Transform}
This section describes the use of the Scattering Transform as an unsupervised representation and as part of hybrid unsupervised learning. First we evaluate the scattering as an unsupervised representation using the CIFAR-10 and ImageNet datasets, then we show that it can be used inside common unsupervised learning schemes by proposing a hybrid GAN combined with a Scattering Transform, which synthesizes Scattering Coefficients from random Gaussian noise on $32\times32$ color images from ImageNet. Using the reconstruction proposed in Section \ref{sec:reconst} we show that we can generate images from this GAN model.

\subsection{Scattering as an Unsupervised Representation}
\label{unsup}
We first consider the CIFAR-10 dataset used in Section \ref{sec:cifar} and perform an experiment that allows us to evaluate the scattering transform as an unsupervised representation with a complex non-convolutional classifier. In a second experiment, we consider the linear classification task on ILSVRC 2012 often used to evaluate unsupervised representations \cite{arandjelovic2017look}.

For CIFAR-10, as in Section \ref{sec:cifar}, we used $J=2$ which means the output of the scattering stage will be $8\times8$ spatially and 243 in the channel dimension. This task has been commonly evaluated on CIFAR-10 with a non-linear classifier \cite{oyallon2015deep} and we thus consider the use of a MLP. We follow the training procedure prescribed in \cite{zagoruyko2016wide} utilizing SGD with momentum of 0.9, batch size of 128, weigh decay of $5\times10^{-4}$, and modest data augmentation of the dataset by using random cropping and flipping. The initial learning rate is 0.1, and we reduce it by a factor of 5 at epochs 60, 120 and 160. The models are trained for 200 epochs in total. We used the same optimization and data augmentation pipeline for training and evaluation in both cases. We utilize batch normalization  at all layers which leads to a better conditioning of the optimization \citep{ioffe2015batch}. Table \ref{tab:CIFAR_Main} reports the accuracy  in the unsupervised and supervised settings and compares them to other approaches. Combining the scattering transform with a NN classifier consisting of 3 hidden layers, with width $1.1\times10^4$, we show that one can obtain a new state of the art classification for the case of unsupervised convolutional layers. More numerical comparisons with other unsupervised methods, such as random networks, can be found in \cite{oyallon2015deep}. Scattering based approaches outperform all methods utilizing learned and not-learned unsupervised features, further demonstrating the discriminative power of the scattering network representation.

For the ILSVRC-2012 dataset we use a common evaluation based on training a linear classifier on top of the unsupervised representation \cite{arandjelovic2017look}. We used a standard training protocol with cross-entropy loss on top of a scattering transform produced with $J=4$. We apply standard data augmentation, optimizing with stochastic gradient descent with momentum $0.9$, weight decay set to $1e-7$, and learning rate drops at epochs 20, 40, and 60. The results are shown in Table \ref{tab:unsupervised} \mynotes{and are compared with unsupervised and self-supervised baselines}. Observe that a Scattering Transform improves significantly from a random baseline \cite{arandjelovic2017look}, and that it recognize a large number of images even when only considering the top result. The accuracy of a random baseline is still high, because the small support of the convolutional operators already incorporates some geometric structures in this type of pipeline. Modern learned unsupervised representations however can improve on this result. 
%In the case of the Scattering Transform, as no adaptation to the specific non-geometric bias is performed, it can not outperform methods such as Fisher Vectors combined with SIFT  \cite{sanchez2011high}.

\begin{table}
\begin{center}
\begin{tabular}{|l|c|c|c|}
\hline
\bf Method  &\bf Top 1   \\
\hline

Scattering + 1 FC  &17.4 $\%$  \\% &44.4  & 21.6 \\
Random CNN \cite{arandjelovic2017look}&12.9 $\%$ \\% &44.4  & 21.6 \\
\hline
Pathak et al \cite{pathak2016context} & 22.3$\%$ \\
Doersch et al \cite{doersch2015unsupervised} & 31.7$\%$ \\
\begin{edit}Donahue et al \cite{donahue2016adversarial} \end{edit}&31.0$\%$ \\
Noroozi and Favaro et al \cite{noroozi2016unsupervised} & 34.7$\%$ \\
Arandjelovic et al \cite{arandjelovic2017look} &  32.6 $\%$ \\

\hline
\end{tabular}
\end{center}
\caption{Comparison of the top-1 accuracy from unsupervised and \mynotes{ self-supervised} representation, on the ImageNet dataset, evaluated as ours with a linear classifier. We compare to a reported result of a similar architecture and random initialization. We also show result of learned unsupervised representations for reference. \mynotes{Baselines for the linear classification results are taken from \cite{arandjelovic2017look}}.}
\label{tab:unsupervised}
\end{table}

%ADVERSARIAL EXAMPLES FROM A LINEAR MODEL(even if it is described later)
In order to test the robustness of the Scattering Network w.r.t.\ adversarial examples, we used the simple sign gradient attack~\cite{goodfellow2014adversarial}. We build adversarial examples that fool our linear layer, which means for a given $x$ classified as $c$ that we desire to force the classifer to erroneously classify as $\tilde c\neq c$, we find the smallest $\epsilon_x$ such that:
\[\text{class}(x+\epsilon_x)= \tilde c.\]

\begin{figure}%[h]
\begin{center}
  \subcaptionbox{Original image $x$, well classified with output probability: \texttt{0.35, tiger cat}}{\includegraphics[width=0.30\columnwidth]{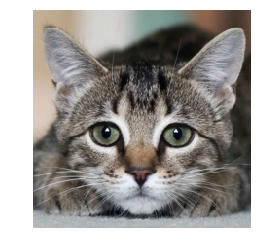}}
  \subcaptionbox{Adversarial sample $\tilde x$, wrongly classified with output probability with $\epsilon=0.15$: \texttt{0.02, magnetic compass}}{\includegraphics[width=0.30\columnwidth]{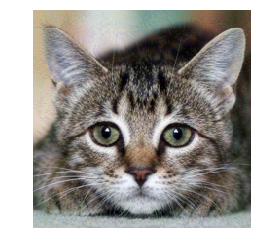}}
  \subcaptionbox{$x-\tilde x$ (magnified for a better visualization)}{\includegraphics[width=0.30\columnwidth]{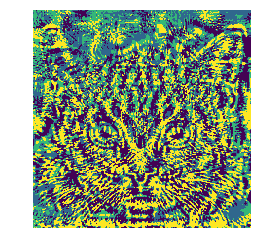}}\\
\end{center}
\caption{Adversarial examples obtained from a Scattering Transform followed by a linear classifier on ImageNet.}\label{adversarial}
\end{figure}

In our case, candidates for $\epsilon$ are given by vectors collinear to the gradient sign in the direction of $\tilde c$ as explained in \cite{goodfellow2014explaining}. Results are shown in Figure \ref{adversarial}.\begin{edit}It shows that being only 1-Lipschitz is not sufficient to be visually robust to such artifacts, when combined only with a linear classifier; using non-linear classifier, such as a CNN, designed to be robust to predefined noises could permit to tackle this issue.\end{edit}%In particular, even with a  mathematically well defined scattering transform which is stable to additive noise, it is possible to build visually indistinguishable images that fool an even simple linear classifier.

\subsection{Hybrid Unsupervised Learning with Scattering GAN}\label{sec:HybridUnsupLearnScatteringGAN}
In this section we propose to construct a Generative Adversarial Network (GAN) in the space of scattering coefficients. This essentially constructs a hybrid generator and discriminator. The GAN is a state-of-the-art generative modeling framework. The use of the learned generator on top of a scattering transform can be well motivated if we consider the scattering transform as good a model of low level texture \cite{bruna2013invariant}. Furthermore, as extensive data augmentation is often not required, it is possible to store scattering representations that have a smaller spatial resolution, permitting us to try rapidly a variety of architectures. We demonstrate in the following that a scattering representation can be used as the initialization of a generative model, similar to the classification case.

We follow the Deep Convolutional Generative Adversarial Network architectures proposed in \cite{radford2015unsupervised} in order to generate signals in the scattering space. We consider color images from the resized ImageNet dataset of size $32\times32$ in the $YUV$ space that are processed by a Scattering Transform with $J=2$. The scattering coefficients were renormalized to lie between $-1$ and $1$. Their scattering representations are then fed to the generator and discriminators of our Scattering-DCGAN. In particular, the generator aims to synthesize scattering coefficients from a Gaussian noise with $d=100$. They are represented in Table  \ref{tab:archigan}. Moreover we apply the recently proposed Wasserstein distance based objective \cite{gulrajani2017improved,arjovsky2017wasserstein}.

We now describe our training procedure. We run the Adam optimizer for both the discriminator and generator during 600k iterations without observing significant instabilities during the optimization. The discriminator is trained  during 5 successive iterations and the generator only 1, as done in \cite{gulrajani2017improved}, because we observed it leads to more realistic images. The generator takes as input a latent variable of 100 dimensions.

\begin{figure}%[h]
\begin{center}
%\framebox[4.0in]{$\;$}

\includegraphics[width=7cm]{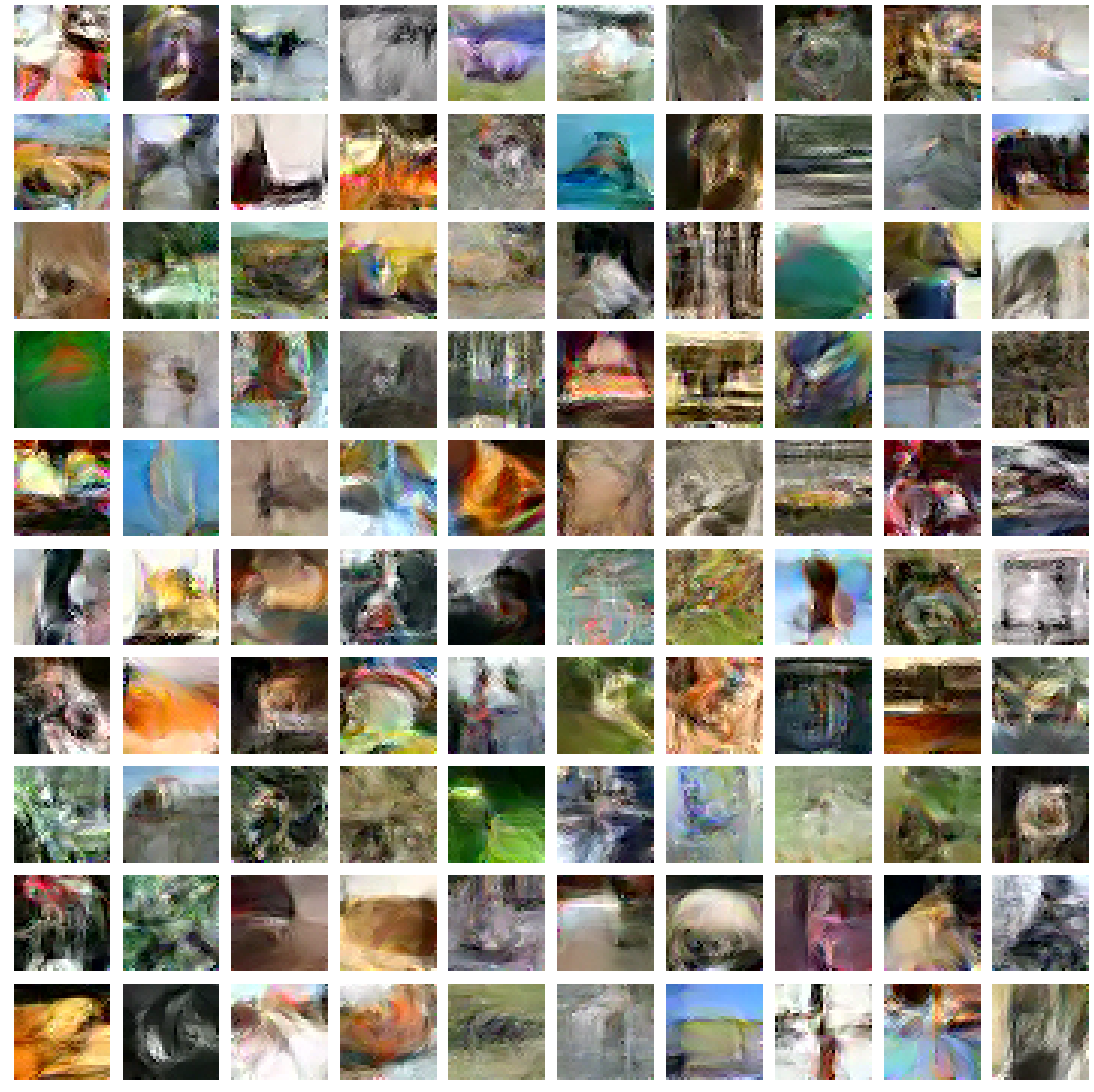}
\end{center}
\caption{Samples generated by the Scattering-DCGAN.  See Section \ref{sec:HybridUnsupLearnScatteringGAN} for details.}\label{samplesgan}
\end{figure}

Section \ref{sec:reconst} shows that the scattering transform can be used to reconstruct images. We thus recover images generated from our model from the generated scattering coefficients, and they are shown in Figure \ref{samplesgan}. These images are qualitatively similar to other baselines, and it shows how one can use the scattering transform with more complex models. Generating coherent Scattering coefficients that leads to real images is  challenging: the non-surjectivity of the scattering transform is due to physical constraints(e.g. interactions between different coefficients), yet we however did not incorporate this knowledge in our architectures.

%\todo{We need to justify how its non-trivial that scattering would generate coherent images}

\begin{table}
    \centering
    \begin{tabular}{c|c}
    \textbf{Generator}&       \\\hline
    random uniform & Input size 100 \\\hline
    2x2 Trans. Conv.     & stride 1, batch norm, LeakyReLU, 256 out \\\hline
    4x4 Trans. Conv. & stride 2, pad 1,batchnorm, LeakyReLU,128 out \\\hline
    4x4 Trans. Conv. & stride 2, pad 1,batchnorm, tanh, 243x8x8 out  \\\hline
      \textbf{Discriminator}&       \\\hline
    random uniform & Input size 243x8x8 \\\hline
    4x4 Conv.     & stride 1, batchnorm, LeakyReLU, 128 out \\\hline
    4x4 Conv. & stride 2, pad 1,batchnorm, LeakyReLU,256 out \\\hline
    4x4 Conv. & stride 2, pad 1, LeakyReLU, 256 out  \\\hline
    1x1 Conv. & stride 1,batchnorm, 256  out  \\\hline
    Fully connected layer &  \\\hline
    \end{tabular}
\caption{Architecture of the Discriminator and Generator of the Scattering-DCGAN.}
\label{tab:archigan}
\end{table}

\section{Learning Scattering}
\addition

Many theoretical arguments of deep learning rely on the universal approximation theorem \citep{cybenko1989approximation}. The flexibility of this deep learning frameworks raises the following question: can we approximate the first scattering layers by a deep network?

In order to explore this question, we consider a 5-layer convnet as a candidate to replace our scattering network on CIFAR10. Its architecture is described in Table \ref{tab:scatapprox}, and it has the same output size as a scattering network. It has two downsampling steps, in order to mimic the behavior of a scattering network with $J=2$.  We build a hybrid architecture, i.e.\ scattering followed by a Cascaded CNN, described in Table \ref{tab:scatapprox} that leads to 91.4\% on CIFAR10.  Then we replace the scattering part by the CNN of Table \ref{tab:scatapprox}, i.e.\ the Scattering Approximator. We train it, keeping the weights of the Cascaded CNN layers constant and equal to the optimal solution found with the scattering. Instead of minimizing a loss between the output of a scattering network and this network, we  target the best input for the fixed convnet given the classification task.

This architecture can achieve $1\%$ accuracy below the original pipeline, which indicates it is possible to learn the Scattering representation. Using a shallower network seems to degrade the performances, but we did not investigate this question further. In any case, the learned network will not have any guarantee of stability properties present in the original scattering transform.

\begin{table}

    \centering
    \begin{tabular}{c|c}
    \textbf{Scattering approximator}&       \\\hline
    3x3  Convolution     & stride 1, batch norm, ReLU, 128 output \\\hline
        3x3  Convolution     & stride 2, batch norm, ReLU, 128 output \\\hline
3x3  Convolution     & stride 1, batch norm, ReLU, 128 output \\\hline
3x3  Convolution     & stride 1, batch norm, ReLU, 256 output \\\hline
3x3  Convolution     & stride 2, batch norm, ReLU, 243 output \\\hline

      \textbf{Cascaded CNN}&       \\\hline
    3x3  Convolution      $\times 10$& stride 1, batch norm, ReLU, 128 output \\\hline
        3x3  Convolution    $\times 10$ & stride 1, batch norm, ReLU, 256 output  \\\hline

    Averaging layer &  \\\hline
    Fully connected layer &  \\\hline
    \end{tabular}
\caption{Architecture of the Scattering approximator.}
\label{tab:scatapprox}
\end{table}

\section{Conclusion}
\input{conclusion}
\IEEEpeerreviewmaketitle

%We would like to thank Mathieu Andreux,   St\'{e}phane Mallat,  for helpful discussions. 

\section*{Acknowledgment}
The authors would like to thank  Mathieu Andreux, Tom\'{a}s Angles, Joan Bruna, Carmine Cella, Bogdan Cirstea, Michael Eickenberg, St\'ephane Mallat, Louis Thiry for helpful discussions and support. 
The authors would also like to thank Rafael Marini and Nikos Paragios for use of computing resources. We would like to thank Florent Perronnin for providing important details of their work. This work is funded by the ERC grant InvariantClass 320959, via a grant for PhD Students of the Conseil r\'egional d'Ile-de-France (RDM-IdF), Internal Funds KU Leuven, FP7-MC-CIG 334380, DIGITEO 2013-0788D - SOPRANO, NSERC Discovery Grant RGPIN-2017-06936, an Amazon Research Award to Matthew Blaschko, and by the Research Foundation - Flanders (FWO) through project number G0A2716N. We thank also the CVN (CentraleSupelec) for providing financial support.

\ifCLASSOPTIONcaptionsoff
  \newpage
\fi



%\bibliographystyle{IEEEtran}

\bibliographystyle{ieee}
\bibliography{iccv}

%\begin{thebibliography}{1}

%\end{thebibliography}

%\begin{biography}[{\includegraphics[width=1in,height=1.25in,clip,keepaspectratio]{mshell}}]{Michael Shell}

\vspace{-14 mm}
%\begin{IEEEbiography}[{\includegraphics[width=1.15cm,height=1.15cm,clip,keepaspectratio]{edouardo}}]{Edouard Oyallon}
\begin{IEEEbiography}[\IEEEbiographynophoto{}]{Edouard Oyallon}
is a faculty member at CentraleSupelec, University of Paris-Saclay. He completed his PhD at the Ecole Normale Sup\'{e}rieure (ENS) in 2017.
\end{IEEEbiography}

% He completed his PhD at the Ecole Normale Supérieure.%  working with Stephane Mallat 

\vspace{-14mm}

\begin{IEEEbiography}[\IEEEbiographynophoto{}]{Sergey Zagoruyko}%[{\includegraphics[width=1.15cm,height=1.25cm,clip,keepaspectratio]{zagoruys}}]{Sergey Zagoryuko}
is a PhD student at Ecole des Ponts ParisTech. 
\end{IEEEbiography}

\vspace{-14 mm}

%\begin{IEEEbiography}[{\includegraphics[width=1.15cm,height=1.25cm,clip,keepaspectratio]{gabrielhuang}}]{Gabriel Huang}
\begin{IEEEbiography}[\IEEEbiographynophoto{}]{Gabriel Huang}
is a PhD student at MILA and DIRO at the University of Montreal.
\end{IEEEbiography}
\vspace{-14 mm}

%\begin{IEEEbiography}[{\includegraphics[width=1.25cm,height=1.25cm,clip,keepaspectratio]{komodakis}}]{Nikos
%Komodakis}
\begin{IEEEbiography}[\IEEEbiographynophoto{}]{Nikos Komodakis}
is a faculty member at Ecole des Ponts ParisTech.
\end{IEEEbiography}
\vspace{-14 mm}

%\begin{IEEEbiography}[{\includegraphics[width=1.25cm,height=1.25cm,clip,keepaspectratio]{lacoste_julien}}]{Simon Lacoste-Julien}
\begin{IEEEbiography}[\IEEEbiographynophoto{}]{Simon Lacoste-Julien}
is a faculty member at MILA and DIRO at the University of Montreal, and a CIFAR fellow.
\end{IEEEbiography}

\vspace{-14 mm}
%\begin{IEEEbiography}[{\includegraphics[width=1.25cm,height=1.25cm,clip,keepaspectratio]{blaschko}}]{Matthew Blaschko}
\begin{IEEEbiography}[\IEEEbiographynophoto{}]{Matthew Blaschko}
is a faculty member in the Electrical Engineering department at KU Leuven.
\end{IEEEbiography}
\vspace{-14 mm}

%\begin{IEEEbiography}[{\includegraphics[width=1.25cm,height=1.25cm,clip,keepaspectratio]{belilovsky}}]
\begin{IEEEbiography}[\IEEEbiographynophoto{}]{Eugene Belilovsky}
 is a postdoctoral fellow at MILA at the University of Montreal. He completed  a joint PhD at CentraleSupelec and at KU Leuven. \end{IEEEbiography}


%\vfill


\end{document}

%% file: intro_abstract.tex
%%%%%%%%% ABSTRACT
\begin{abstract}

Scattering networks are a class of \textit{designed} Convolutional Neural Networks (CNNs) with fixed weights. We argue they can serve as generic representations for modelling images. In particular, by working in scattering space, we achieve competitive results both for supervised and unsupervised learning tasks\mynotes{, while making progress towards constructing more interpretable CNNs.} 
For supervised learning, we demonstrate that the early layers of  CNNs do not necessarily need to be learned, and can be replaced with a scattering network instead. 
Indeed, using  hybrid architectures, we achieve the best results with predefined representations to-date, while being competitive with end-to-end learned CNNs. Specifically, even applying a shallow cascade of small-windowed scattering coefficients followed by $1\times1$-convolutions results in AlexNet accuracy on the  ILSVRC2012 classification task. Moreover, by combining scattering networks with deep residual networks, we achieve a single-crop top-5 error of $11.4\%$ on ILSVRC2012. Also, we show they can yield excellent performance in the small sample regime on CIFAR-10 and STL-10 datasets, exceeding their end-to-end counterparts, through their ability to incorporate geometrical priors.
For unsupervised learning, scattering coefficients can be a competitive  representation that permits image recovery. We use this fact to train hybrid GANs to generate images. 
Finally, we empirically analyze several properties related to stability and reconstruction of images from scattering coefficients.
\end{abstract}

\begin{IEEEkeywords}
Scattering transform, Wavelets, Deep neural networks, Invariance.
\end{IEEEkeywords}

%%%%%%%%% BODY TEXT

\section{Introduction}

\IEEEPARstart{N}{atural} image processing tasks are  high dimensional problems that require introducing lower dimensional representations: in the case of image classification, they must reduce the non-informative image variabilities, whereas for image generation, it is desirable to parametrize them. For example, some of the main source of variability are often due to geometrical operations such as translations and rotations. Then, an efficient classification pipeline necessarily builds invariants to these variabilities, whereas mapping to those sources of variabilities is desirable in the context of image generation. Deep architectures build representations that lead to state-of-the-art results on image classification tasks \citep{he2015deep}. These architectures are designed as very deep cascades of non-linear end-to-end learned modules \citep{lecun2010convolutional}.  When trained on large-scale datasets they have been shown to produce representations that are transferable to other datasets \citep{zeiler2014visualizing,huh2016makes}, which indicates they have captured generic properties of a supervised task that consequently do not need to be learned. Indeed several works indicate geometrical structures in the filters of the earlier layers  of Deep CNNs \cite{krizhevsky2012imagenet,waldspurger2015these}. However, understanding the precise operations performed by those early layers is a complicated \cite{szegedy2013intriguing,oyallon2017building} and possibly intractable task. In this work we investigate if it is possible to replace these early layers by simpler cascades of non-learned operators that reduce and parametrize variability while retaining all the discriminative information.

Indeed, there can be several advantages to incorporating pre-defined geometric priors via a hybrid approach of combining pre-defined and learned representations. First, end-to-end pipelines can be data hungry and ineffective when the number of samples is low. Secondly, it could lead to more interpretable classification pipelines, which are amenable to analysis, and permits the performance of parallel transport along the Euclidean group. Finally, it can reduce the spatial dimensions and the required depth of the learned modules, improving their computational and memory requirements. 

A potential candidate for an image representation is the SIFT descriptor \cite{lowe1999object}, which was widely used before 2012 as a feature extractor in classification pipelines \cite{sanchez2011high,sanchez2013image}. This representation was typically encoded via an unsupervised Fisher Vector (FV) and fed to a linear SVM. However, several works indicate that this is not a generic enough representation on top of which to build further modules  \cite{le2011learning,bo2013multipath}. Indeed end-to-end learned features produce substantially better classification accuracy.

A Scattering Transform \cite{mallat2012group,bruna2013invariant,sifre2013rotation} is an alternative that solves some of the issues with SIFT and other predefined descriptors. In this work, we show that contrary to other proposed descriptors \cite{tola2010daisy}, a Scattering Network can avoid discarding information. Indeed, a Scattering Transform is not quantized, and the loss of information is avoided thanks to a combination of wavelets and non linear operators. Furthermore, it is shown in \citep{oyallon2015deep} that a Scattering Network provides a substantial improvement in classification accuracy over SIFT. A Scattering Transform also provides certain mathematical guarantees, which CNNs generally lack. Finally, wavelets are often observed in the initial layers, as in the case of AlexNet \cite{krizhevsky2012imagenet}. Thus, combing the two approaches is natural.

This article is an extended version of \cite{oyallon:hal-01495734}. Our main contributions are as follows. First, we design and develop a fast algorithm to compute a Scattering Transform to use in a deep learning context. We demonstrate that using supervised local descriptors obtained by shallow $1\times1$ convolutions with very small spatial window sizes obtains AlexNet accuracy on the ImageNet classification task (Subsection \ref{supervised_encoding}). We show empirically that these encoders build explicit invariance to local rotations (Subsection \ref{sec:sle_first_sec}). Second, we propose hybrid networks that combine scattering with modern CNNs  (Section \ref{small2}) and show that using scattering and a ResNet of reduced depth, we obtain similar accuracy to ResNet-18 on ImageNet (Subsection \ref{hybimnet}). Then, we study adversarial examples to the Scattering Transform with a linear classifier. We then develop a procedure to reconstruct an image from its scattering representation in Section~\ref{sec:reconst} and show that this can be used to incorporate the scattering transform in a hybrid Generativel Adversarial Network in Section~\ref{sec:HybridUnsupLearnScatteringGAN}. Finally, we demonstrate in Subsection \ref{verysmall} that scattering permits a substantial improvement in accuracy in the setting of limited data.

Our highly efficient GPU implementation of the scattering transform is, to our knowledge, orders of magnitude faster than any other implementations, and allows training very deep networks while applying scattering on the fly. Our scattering implementation\footnote{\url{http://github.com/edouardoyallon/pyscatwave}} and pre-trained hybrid models\footnote{\url{http://github.com/edouardoyallon/scalingscattering}} are publicly available. 

\begin{edit}

\section{Related Work}

Closely related to our work, \cite{perronnin2015fisher} proposed a hybrid representation for large scale image recognition combining a predefined representation and Neural Networks (NN), that uses a Fisher Vector (FV) encoding of SIFT and leverages NNs as scalable classifiers. In contrast we use the scattering transform in combination with convolutional architectures and show hybrid results that well exceed those of \cite{perronnin2015fisher}. 

A large body of recent literature has also considered unsupervised and  self-supervised learning for constructing discriminative image features \cite{arandjelovic2017look,donahue2016adversarial} that can be used in subsequent image recognition pipelines. However, to the best of our knowledge on complex datasets such as imagenet these representations do not yet approach the accuracy of supervised methods or hand-crafted unsupervised representations. In particular the FV encoding discussed above is an unsupervised representation that has outperformed any unsupervised learned representation on the imagenet dataset \cite{sanchez2013image}. 

With regards to the algorithmic implementation of the Scattering Transform, former implementations  \cite{bruna2013invariant,anden2014scatnet} were only scaled for CPU as they retain too many intermediate variables, which can be too large for GPU use. A major contribution of our work is to propose an efficient approach which fits in GPU memory, which subsequently allows a much faster computational time than the CPU implementations. This is essential for scaling to the ImageNet dataset.

Concurrent to our work the Scattering Transform was also recently used in a context of generative modeling \cite{angles2018generative}: it is shown that by inverting the scattering transform, it is possible to generate images in a similar fashion as GANs. We however adopt a rather different approach by building hybrid GANs that directly learn to generate Scattering coefficients, which we reconstruct back into images.
\end{edit}

%shows an architecture that can locally encode scattering features before jointly with an NN classifier. In the experimental Section we show how this architecture consisting of only 1x1 convolutions and 2 FC layers can obtain accuracies comparable to that of the AlexNet \cite{}. We analyze the properties of this interpertable hybrid network and show. In Section {} we show that cascading standard convolutional architectures with the scattering network can yield competitive results on Imagenet ILSVRC and CIFAR-10 comparable to those of the original architecutres, learning convolutions only on a smaller spatial resolution input, all the while having stronger theoretical guarantees for its representations. In the setting of limited data we show that we can obtain large improvements in classification performance using the hybrid architectures.

%We explain how we build our scattering network, describe its stability properties and exhibit our learning pipeline. Section \ref{Exp} shows that our network provides competitive results on CIFAR10, CIFAR100 and STL10, while having theoretical guarantees for its representations, in both setting with limited data or not. 
%The experiments can be reproduced using ScatWave %\footnote{Code can be found here: \href{https://github.com/edouardoyallon/scatwave}{https://github.com/edouardoyallon/scatwave}}, an implementation of our algorithm  in Torch, which we make publicly available. More details about the software are available in the Appendix A.

%% file: scattering_network.tex
%Stability properties are discussed in Subsection \ref{stab} and in Apprendix B.

%\subsubsection{A cascade of wavelets and modulus}

%In this section, we recall the definition of the scattering transform. that reduces translation variability while being discriminative. % Again, it is the cascade of wavelet transforms, and modulus non-linearity which are finally spatially averaged. Since a modulus is non-expansive, and a wavelet transform is almost a linear isometry, a scattering transform is also non-expansive. The local averaging of this representation  builds a local invariance to translation. 

%Sx Scattering representation VS ``layer map''
%Sx(u) Scattering coefficient VS ``channels feature map''
%S scat transform VS scattering network

% Scattering diagram for S

% transition between 2.2 and 2.3, we do a cascade, but 2012 before <> after

We now describe the scattering transform and motivate its use as a generic input for supervised tasks. A scattering network belongs to the class of CNNs whose filters are fixed wavelets \citep{oyallon2015deep}. The construction of this network has strong mathematical foundations \citep{mallat2012group}, meaning it is well understood, relies on  few parameters, and is stable to a large class of geometric transformations.  In general, the parameters of a scattering transform do not need to be adapted to the bias of the dataset \citep{oyallon2015deep}, making its output a suitable generic representation.
% gabriel: i changed respresentation to scattering transform because it makes more sense to talk about the parameters of a transformation than of a representation.
 
We then propose and motivate the use of supervised CNNs built on top of the scattering network. Finally we propose supervised encodings of scattering coefficients using 1x1 convolutions, which can retain interpretability and locality properties.

%The former   

% We propose to  encode locally each scattering coefficient, the scattering transform at each position, as well as to perform a  independently each scattering output. We show that the former approach based on local supervised encoding bears both characteristic from techniques in classical computer vision, such as \cite{ref}, and modern deep learning methods used in computer vision \cite{ref}. 

\subsection{The Scattering Transform}
\label{scatnet}

\begin{figure}%[h]
\begin{center}
%\framebox[4.0in]{$\;$}
\label{archi}
%\fbox{\rule[-.5cm]{0cm}{4cm}   

\begin{tikzpicture}%[draw=black!50]
\node at (-1,0) [draw=none,line width=0] (x) {$x$};

\node at (0,0) [minimum width=0.8cm,draw,color=black!60!green] (W1) {$|W_1|$};
\node at (1.5,0) [minimum width=0.8cm,draw,color=red] (W2) {$|W_2|$};
\node at (2.8,0) [minimum width=0.8cm,draw] (A1) {$A_J$};
\node at (3.8,0) [draw=none,line width=0] (S) {$Sx$};

\draw[->,,black] (x) -- (W1);
\draw[->,black] (W1) -- (W2);
\draw[->,black] (W2) -- (A1);
\draw[->,black] (A1) -- (S);
%\draw[->,thick,black] (A1) -- (C0);
%\draw [-<,black] (W1.north)  -| (0,0.6) --(2.8,0.6)|- (2.8,0.32);
\draw [-<,black] (0.8,0)  -| (0.8,0.5) --(2.8,0.5)|- (2.8,0.32);
\draw [->,black] (-0.6,0) -| (-0.6,-0.5) --(2.8,-0.5)|- (2.8,-0.28);

\end{tikzpicture}

\end{center}
\caption{A scattering network. $A_J$ concatenates the averaged signals (cf.\ Section~\ref{scatnet}).%\todo[inline]{MB: define the variables and the semantics of boxes and arrows here}
}\label{fig:ScatteringNetwork}
\end{figure}

In this section, we recall the definition of the scattering transform, introduced in~\cite{bruna2013invariant}, and clarify it by illustrating how to concretely apply it on a discrete image. In general, consider a signal $x(u)$, with $u$ the spatial position index and an integer $J\in \mathbb{N}$, which is the spatial scale of our scattering transform. In particular, when $x$ is a grayscale image, we write $x[p]$ its discretization, where $p_1,p_2\leq N$. Let $\phi_J$ be a local averaging filter with a spatial window of scale $2^J$ (here, a Gaussian smoothing function). 
We obtain the \textit{zeroth order} scattering coefficients $S^0x(u)=A_Jx(u)=x\star \phi_J(2^Ju)$ by applying\footnote{In this work, $\star$ denotes convolution, and has higher precedence than function evaluation.} a local averaging operator $A_J$, followed by an appropriate downsampling of scale $2^J$. The zeroth order scattering transform is approximately invariant to translations smaller than $2^J$, but also results in a loss of high frequencies, which are necessary to discriminate signals. In our grayscale image example, $S^0x$ is a feature map of resolution $\frac{N}{2^J}\times \frac{N}{2^J}$ with a single channel. 

%%In the following, we use Morlet wavelets~\cite{MorletWavelets1982}, which are defined for appropriate $\omega_0,\sigma$ by:
%\[\psi(u)=e^{\frac{\Vert u\Vert^2}{2\sigma^2}}(C-e^{i\omega_0^Tu}) , \] where $C$ is set such that this function has 0 integral.

A solution to avoid the loss of high frequency information is to use wavelets. A wavelet is an integrable function with zero mean, which is localized both in Fourier and space domain~\cite{Mallat1989TMS}.  A family of wavelets is obtained by dilating a complex mother wavelet $\psi$ (here, a Morlet wavelet) such that $\psi_{j,\theta}(u)=\frac 1 {2^{2j}}\psi(r_{-\theta}\frac{u}{2^j})$, where $r_{-\theta}$ is the rotation by $-\theta$, and $j\geq 0$ is the scale of the wavelet. Thus, a given wavelet $\psi_{j,\theta}$ has its energy concentrated at a scale $j$ in the angular sector $\theta$. Let $L\in \mathbb{N}$ be an integer parametrizing a discretization of $[0,2\pi]$. A wavelet transform is the convolution of a signal with the family of wavelets introduced above, followed by an appropriate downsampling:
\begin{equation*}
W_1x(j_1,\theta_1,u)=\{x\star \psi_{j_1,\theta_1}(2^{j_1}u)\}_{j_1\leq J,\theta_1=2\pi\frac{l}{L},1\leq l\leq L}
\end{equation*}
Observe that $j_1$ and $\theta_1$ have been discretized -- the wavelet is chosen to be selective in angle and localized in the Fourier domain. 
%\if false
%thus the sampling is chosen such that $(\theta_1,j_1)\rightarrow W_1x(u,\theta_1,j_1)$ is regular enough. 
%Besides, the wavelet transform has been spatially oversampled by a factor 1.
 %The wavelet parameters and this discretization were already chosen in \citep{oyallon2015deep}, where this representation is shown to be generic, so we have used the same hyper-parameters.
%\fi 
With appropriate discretization \cite{oyallon2015deep}, $\{A_Jx,W_1x\}$ is approximatively an isometry on the set of signals with limited bandwidth, which implies that the energy of the signal is preserved. This operator then belongs to the category of multi-resolution analysis operators, each filter being excited by a specific scale and angle, but with the output coefficients not being invariant to translation. To achieve invariance we cannot apply $A_J$ directly to $W_1x$ since it would result in a trivial invariant, namely zero.

%We thus build the first order scattering coefficients. A

To tackle this issue, we first apply a non-linear point-wise complex modulus to $W_1x$, followed by an averaging $A_J$, and a downsampling of scale $2^J$, which builds a non-trivial invariant. Here, the mother wavelet is analytic, thus $|W_1x|$ is regular \citep{bernstein2013generalized} which implies that the energy of $|W_1x|$ in the Fourier domain is more likely to be contained in a lower frequency regime than $W_1x$. Thus, $A_J$ preserves more energy of $|W_1x|$. It is possible to define \[S^1x=A_J|W_1|x,\] which can also be written as: \[S^1x(j_1,\theta_1,u)=|x\star\psi_{j_1,\theta_1}|\star\phi_J(2^Ju);\] these are the \textit{first-order} scattering coefficients. Following deep-learning terminology, each $S^1x(j_1,\theta_1,\cdot)$ can be thought of as a one channel in a feature map. Again, the use of averaging  builds an invariant to translation up to $2^J$. In our grayscale image example, $S^1x[p]$ is a feature map of resolution $\frac{N}{2^J}\times \frac{N}{2^J}$ with $JL$ channels.

To recover some of the high-frequencies lost due to the averaging applied on the first order coefficients, we apply again a second wavelet transform $W_2$ (with the same filters as $W_1$) to each channel of the first-order scatterings, \textit{before} the averaging step. This leads to the \textit{second-order} scattering coefficients \[S^2x=A_J|W_2||W_1|,\] which can also be written as
\[S^2x(j_1,j_2,\theta_1,\theta_2,u)=||x\star \psi_{j_1,\theta_1}|\star \psi_{j_2,\theta_2}|\star \phi_J(2^Ju).\] We only compute paths of increasing scale ($j_1< j_2$) because non-increasing paths have been shown to bear no energy \citep{bruna2013invariant}. In our grayscale image example, $S^2x[p]$ is a feature map of resolution $\frac{N}{2^J}\times \frac{N}{2^J}$ with $\frac 1 2 J(J-1)L^2$ channels (one per increasing path).

We do not compute higher order scatterings, because their energy is negligible \citep{bruna2013invariant}. We call $Sx(u)$(or $S_Jx(u)$) the final scattering coefficient corresponding to the concatenation of the order 0, 1 and 2 scattering coefficients, intentionally omitting the path index of each representation.  A schematic diagram is shown in Figure~\ref{fig:ScatteringNetwork}. In the case of color images, we apply independently a scattering transform to each RGB channel of the image, which means  $Sx(u)$ is a feature map with $3\times \big(1+JL+\frac 1 2 J(J-1)L^2\big)$ channels, and the original image is down-sampled by a factor $2^J$ \cite{bruna2013invariant}. % \textcolor{red}{I moved this from above it fits better here: In this work, we only consider a second order scattering network, on the group of translations.}

This representation has been proved to linearize small deformations of images \cite{mallat2012group}, be non-expansive and almost complete \cite{dokmanic2016inverse,bruna2013audio}, which makes it an ideal input to a deep network algorithm, which can build invariants to this local variability via a first linear operator. We discuss its use as an  initialization of a deep network in the next sections.

\subsection{Efficient Implementation of Scattering Transforms}
The implementation of a Scattering Network must be re-thought to benefit from GPU acceleration. Indeed, a GPU is a device which has a  limited memory size in comparison with a CPU, and thus it is not possible to store intermediate computations. In this section, we show how to solve this problem of memory. We first describe the naive tree implementation \cite{bruna2013invariant,oyallon2015deep} and then our efficient GPU based implementation.

\subsubsection{Tree implementation of computations}

%\addition
\begin{figure}%[h]
\begin{center}
%\framebox[4.0in]{$\;$}
%\fbox{\rule[-.5cm]{0cm}{4cm}   
\includegraphics[width=8cm]{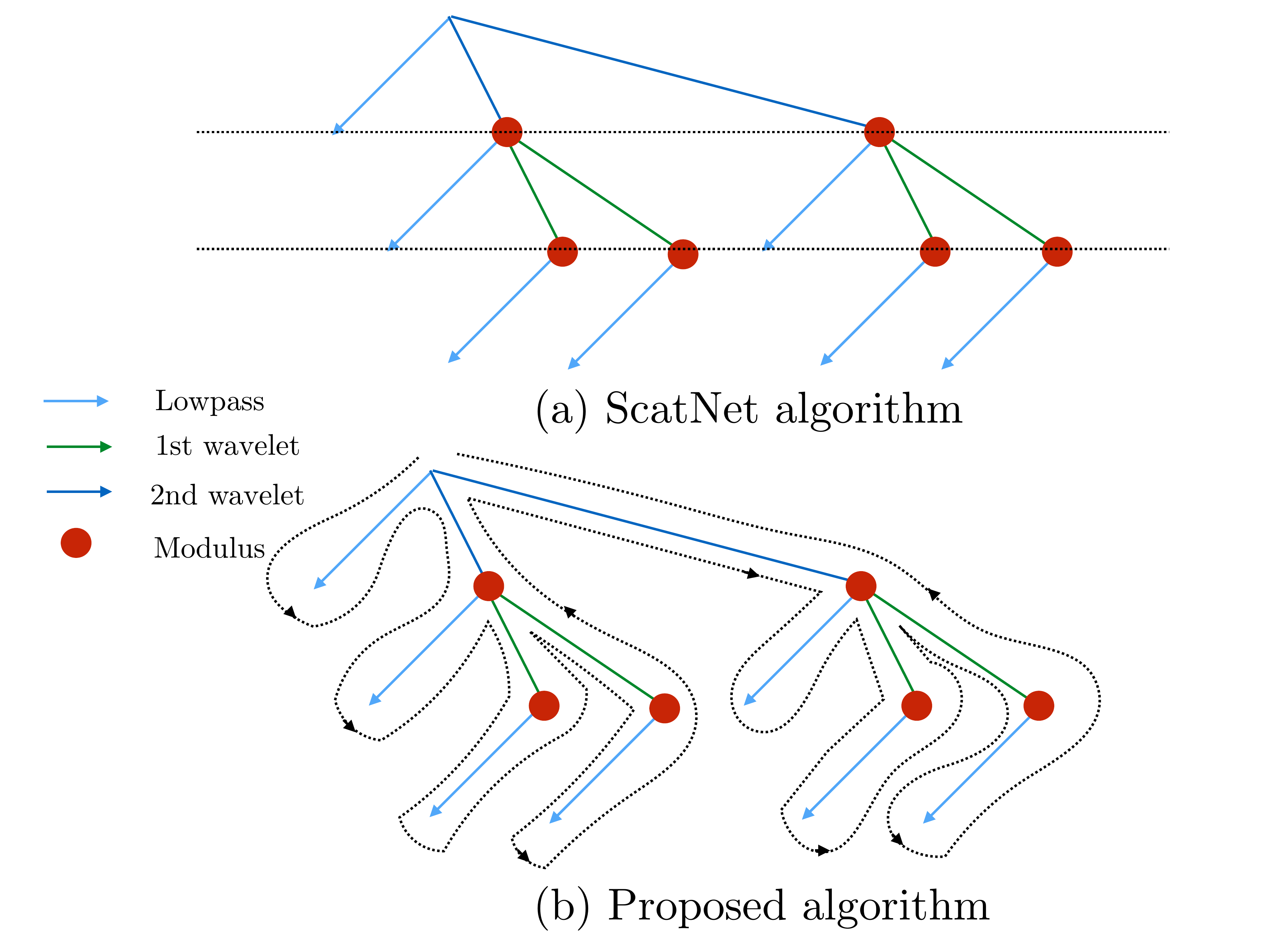}

\end{center}
\caption{Trees of computations for a Scattering Transform. (a) corresponds to the traversal used in the ScatNet software package and (b) to our current implementation (PyScatWave). }
\label{fig:impl}
\end{figure}

We recall the algorithm to compute a Scattering Transform and its implementation in \cite{bruna2013invariant, anden2014scatnet} for order 2 Scattering with a scale of $J$ and $L$ different orientations for the wavelets. We explicitly show this algorithm is not appropriate to be scaled on a GPU. It corresponds to a level order traversal of the tree of computations of the Figure \ref{fig:impl}(a). Let us consider again a discretized input signal $x[p]$ of size $N^{2}$ which is a power of 2, and a spatial sampling of 1.   For the sake of simplicity, we assume that an algorithm such as a symmetric padding has already been applied to $x$ in order to avoid boundary effects that are inherent to periodic convolutions. The filter bank corresponds to $JL+1$ filters: %, e.g.:
\[\{\psi_{\theta,j},\phi_{J}\}_{\theta,j\leq J}.\] We only consider periodized filters, e.g.:
\[\tilde{\psi}_{\theta,j}(u)=\sum_{k_{1},k_{2}}\psi_{\theta,j}(u+(Nk_{1},Nk_{2})) .\]
A first wavelet transform must be applied on the input signal of the Scattering Transform. To this end, (a) a FFT  of size $N$ is applied. Then, (b) $JL$ dot-wise multiplications  with the resulting signal must be applied using the filters in the Fourier domain, $\{\hat{\tilde{\psi}}_{\theta,l}(\omega),\hat{\tilde{\phi}}_{J}(\omega)\}$. Each of the the resulting filtered $x\star\tilde{\psi}_{j,\theta}[p]$ or $x\star\tilde{\phi}_{J}[p]$ signals must be down-sampled by a factor of $2^{j}$ or $2^{J}$, respectively, in order to reduce the computational complexity of the next operations. This is performed by (c) a periodization  of the signal in the Fourier domain, which is equivalent to a down-sampling in the spatial domain, i.e.\ the resulting signal is $x\star\tilde{\psi}_{j,\theta}[2^{j}p]$ or $x\star\tilde{\phi}_{J}[2^{J}p]$. This last operation will lead to an aliasing, because there is a loss of information that can not be exactly recovered with Morlet filters. %\todo{IT IS DONE BELOWneed to introduce morlet wavelets}, because they introduce an aliasing. 
(a') An iFFT is then applied to each of the resulting filtered signals, which are of size $\frac{N^{2}}{2^{j}},j\leq J$. (d) A modulus operator is applied to each of the signals, except to the low pass filter because it is a Gaussian. 
%real filters given\todo{MB: "real filters given" unclear.}. 
The set of filters to be reused at the next layer is $\{|x\star\tilde{\psi}_{j_{1},\theta_{1}}[2^{j_{1}}p]|\}_{\theta_{1}\leq L,j_{1}<J,p_{1}\leq\frac{N}{2^{j_{1}}},p_{2}\leq\frac{N}{2^{j_{1}}}}$ plus a low pass filter.
%\begin{edit}TOREMOVE?- This requires the storing of $\mathcal{O}^{1}$ intermediate coefficients, where:

%\begin{align*}
%\mathcal{O}^{1}&=L\sum_{j_{1}=0}^{J-1}\frac{N^{2}}{2^{2j_{1}}}+\frac{N^{2}}{2^{2J}}\\
%&=N^{2}[L(4\frac{1-4^{-J}}{3})+4^{-J}]  .
%\end{align*}
%This exact step is iterated one more time, on each of the $JL$ wavelet modulus signals, while only considering increasing paths. This means a wavelet transform and a modulus applied on a signal $|x\star\tilde{\psi}_{j_{1},\theta_{1}}[2^{j}p]|$ leads to additional coefficients:
%\begin{align*}
%\mathcal{O}_{j_{1}}^{2}	&=	L\sum_{j_{2}=j_{1}+1}^{J-1}\frac{N^{2}}{2^{2j_{2}}}+\frac{N^{2}}{2^{2J}}\\
%	&=	N^{2}[L(4\frac{4^{-j_{1}-1}-4^{-J}}{3})+4^{-J}]
%	\end{align*}
%Consequently, the total number of coefficients stored for the second order scattering is:
%\begin{align*}
%\mathcal{O}^{2}	&=	%\sum_{j_{1}=0}^{J-1}L\mathcal{O}_{j_{1}}^{2}\\
%	&=	\sum_{j_{1}=0}^{J-1}LN^{2}[L(4\frac{4^{-j_{1}-1}-4^{-J}}{3})+4^{-J}]\\
%	\end{align*}
%Finally, an averaging is applied on the second order wavelet modulus coefficients, which leads to:
%\[\mathcal{O}^{3}=\frac{J(J-1)}{2}L^{2}\frac{N^{2}}{2^{2J}}\]
%additional coefficients.
%In total, the total number of coefficients stored is:
%\[\mathcal{O}=\mathcal{O}^{1}+\mathcal{O}^{2}+\mathcal{O}^{3}\]
%\end{edit}
\mynotes{ This requires the storage of $\mathcal{O}_{j_i,j_p}=L\sum_{j=j_i}^{J-1}\frac{N^{2}}{2^{2j_{p}}}+\frac{N^{2}}{2^{2J}}$ intermediate coefficients for the first layer, where $j_p=j_1,j_i=0$. This step is iterated one more time, on each of the $JL$ wavelet modulus signals, while only considering increasing paths. This means a wavelet transform and a modulus applied on a signal $|x\star\tilde{\psi}_{j_{1},\theta_{1}}[2^{j}p]|$ lead to an additional storage requirement of $\mathcal{O}_{j_{1},j_{2}}$.
Consequently, the total number of coefficients stored for the second layer of the transform is $	\sum_{j_{1}=0}^{J-1}L\mathcal{O}_{j_{1},j_{2}}^{2}$
Finally, an averaging is applied on the second order wavelet modulus coefficients, which leads to a memory usage of $\frac{J(J-1)}{2}L^{2}\frac{N^{2}}{2^{2J}}$
additional coefficients.
Thus in total, the tree implementation requires a storage size of \[\mathcal{O}_{0,j_1}+\sum_{j_{1}=0}^{J-1}L\mathcal{O}_{j_{1},j_2}^{2}+\frac{J(J-1)}{2}L^{2}\frac{N^{2}}{2^{2J}}\]
}
\mynotes{The above approach is far too memory-consuming for a GPU implementation.} For example, for $J=2,3,4,L=8,$ and $N=256$, which corresponds to the setting used on our ImageNet experiments, we numerically have approximately $2M,2.5M,2.6M$ parameters for a single tensor. A parameter is about 4 bytes, thus an image is about 8MB in the smallest case. In the case of batches of size 256 with color images, we thus need at least 6GB of memory simply to store the intermediate tensors used by the scattering, which does not take in account extra-memory used by libraries such as cuFFT for example. In particular, this reasoning demonstrates that a typical GPU with 12GB of memory can not efficiently process images in parallel with this naive approach.

\subsubsection{Memory efficient implementation on GPUs}
We now describe a GPU implementation which tries to minimize the memory usage during the computations. The procedures (a/a'), (b), (c) and (d) of the previous section can be efficiently implemented entirely on GPUs. They are fast, and can be implemented in batches, which permits parallel computations of the scattering representation. This is necessary for  deep learning pipelines, which commonly use batches of data augmented samples. %However, a GPU is a device with a limited amount of memory.

\begin{table}%[h]

\begin{center}
\begin{tabular}{lc|clc|c|}
\bf Input size  & \bf J & \bf ScatNetLight (in s) & \bf PyScatWave (in s)
\\ \hline \\
$32\times 32 \times 3 \times 128$         &2&2.5 & 0.03 \\
$32\times 32 \times 3 \times 128$         &4&13 & 0.20 \\
$128\times 128 \times 3 \times 128$         &2&16 & 0.26 \\
$128\times 128 \times 3 \times 128$         &4&52 & 0.54 \\
$256\times 256 \times 3 \times 128$         &2&160 & 0.71 \\
$256\times 256 \times 3 \times 128$         &3&    & 1.52 \\
$256\times 256 \times 3 \times 128$         &4&    & 1.73 \\
\end{tabular}
\end{center}

%\fbox{\rule[-.5cm]{0cm}{4cm}   
\caption{Comparison of the computation time of a Scattering Transform on CPU/GPU.  PyScatWave (Algorithm~\ref{algo}) significantly improves performance in practice.}\label{speed}
\end{table}

\if false
\begin{algorithm}%[H]
\caption{Pseudo-code of the algorithm used in PyScatWave.}
\begin{algorithmic}[1]
\Function{Scattering}{$x,J$}\Comment{Where x - image, J - scale}
\State $\tilde{U}_{0}^{0}=FFT(x)$
\State $\tilde{U}_{0}^{1}=\hat{\tilde{\phi}}_{J}\odot\tilde{U}_{0}^{0}$
\State $\tilde{U}_{J}^{1}=\text{periodize}(\tilde{U}_{0}^{1},J)$
\State $S_{J}^{0}x=iFFT(\tilde{U}_{J}^{1})$
\For {$\lambda_{1} = (j_{1},\theta_{1})$ to ${(j_{N_1},\theta_{N_1})}$}
    \State $\tilde{U}_{0}^{1}=\hat{\tilde{\psi}}_{\lambda_{1}}\odot\tilde{U}_{0}^{0}$
    \State $\tilde{U}_{j_{1}}^{1}=\text{periodize}(\tilde{U}_{0}^{1},j_{1})$
    \State $\tilde{U}_{j_{1}}^{1}=iFFT(\tilde{U}_{j_{1}}^{1})$
    \State $\tilde{U}_{j_{1}}^{1}=|\tilde{U}_{j_{1}}^{1}|$
    \State $\tilde{U}_{j_{1}}^{1}=FFT(\tilde{U}_{j_{1}}^{1})$
    \State $\tilde{U}_{j_{1}}^{2}=\hat{\tilde{\phi}}_{J}\odot\tilde{U}_{j_{1}}^{1}$
    \State $\tilde{U}_{J}^{2}=\text{periodize}(\tilde{U}_{j_{1}}^{2},J)$
    \State $S_{J}^{1}x[\lambda_{1}]=iFFT(\tilde{U}_{J}^{2})$
    \For{ $\lambda_{2} = (j_{2},\theta_{2})$ to $(j_{N_2},\theta_{N_2})$}
       \State $ \tilde{U}_{j_{2}}^{2}=\hat{\tilde{\psi}}_{\lambda_{2}}\odot\tilde{U}_{j_{1}}^{1}$
       \State $\tilde{U}_{j_{2}}^{2}=\text{periodize}(\tilde{U}_{j_{2}}^{2},j_{2})$
       \State$\tilde{U}_{j_{2}}^{2}=iFFT(\tilde{U}_{j_{2}}^{2})$
       \State $\tilde{U}_{j_{2}}^{2}=|\tilde{U}_{j_{2}}^{2}|$
       \State $\tilde{U}_{j_{2}}^{2}=FFT(\tilde{U}_{j_{2}}^{2})$
       \State $\tilde{U}_{j_{2}}^{2}=\hat{\tilde{\phi}}_{J}\odot\tilde{U}_{j_{2}}^{1}$
       \State $\tilde{U}_{J}^{2}=\text{periodize}(\tilde{U}_{j_{2}}^{2},J)$
       \State $S_{J}^{2}x[\lambda_{1},\lambda_{2}]=iFFT(\tilde{U}_{J}^{2})$
    \EndFor
\EndFor
output: Sx=\{S_{J}^{0}x,S_{J}^{1}x,S_{J}^{2}x\}
\EndFunction
\label{algo}
\end{algorithmic}
\end{algorithm}
\fi

\begin{algorithm}
\begin{edit}

    \SetKwInOut{Input}{Input}
    \SetKwInOut{Output}{Output}

    \underline{function Scattering} $(x,J)$\;
    \Input{Where $x$ - image, $J$ - scale}
    \Output{$scattering(x,J)$}
    $\tilde{U}_{0}^{0}=FFT(x)$\;
    $\tilde{U}_{0}^{1}=\hat{\tilde{\phi}}_{J}\odot\tilde{U}_{0}^{0}$\;
    %$\tilde{U}_{J}^{1}=\text{periodize}(\tilde{U}_{0}^{1},J)$\;
    $S_{J}^{0}x=iFFT(\text{periodize}(\tilde{U}_{0}^{1},J))$\;
    \For{$\lambda_{1} = (j_{1},\theta_{1})$} %to ${(j_{N_1},\theta_{N_1})}$}
      {
      %  $\tilde{U}_{0}^{1}=\hat{\tilde{\psi}}_{\lambda_{1}}\odot\tilde{U}_{0}^{0}$\;
      $\tilde{U}_0^1=\hat{\tilde{\psi}}_{\lambda_{1}}\odot\tilde{U}_{0}^{0}$\;
        $\tilde{U}_{j_{1}}^{1}=FFT(|iFFT(\text{periodize}(\tilde{U}_0^1,j_{1})))$\;
      %  $\tilde{U}_{j_{1}}^{1}=|iFFT(\tilde{U}_{j_{1}}^{1})|$\;
       % $\tilde{U}_{j_{1}}^{1}=|\tilde{U}_{j_{1}}^{1}|$\;
        %$\tilde{U}_{j_{1}}^{1}=FFT(|iFFT(\tilde{U}_{j_{1}}^{1})|)$\;
        $\tilde{U}_{j_{1}}^{2}=\hat{\tilde{\phi}}_{J}\odot\tilde{U}_{j_{1}}^{1}$\;
      % $\tilde{U}_{J}^{2}=\text{periodize}(\tilde{U}_{j_{1}}^{2},J)$\;
        $S_{J}^{1}x[\lambda_{1}]=iFFT(\text{periodize}(\tilde{U}_{j_{1}}^{2},J))$\;
        \For{ $\lambda_{2} = (j_{2},\theta_{2})$, $j_1<j_2$}% to $(j_{N_2},\theta_{N_2})$}
        {
           $ \tilde{U}_{j_{1}}^{2}=\hat{\tilde{\psi}}_{\lambda_{2}}\odot\tilde{U}_{j_{1}}^{1}$\;
           %$\tilde{U}_{j_{2}}^{2}=\text{periodize}(\tilde{U}_{j_{2}}^{2},j_{2})$\;
         %  $\tilde{U}_{j_{2}}^{2}=iFFT(\text{periodize}(\tilde{U}_{j_{2}}^{2},j_{2}))$\;
           $\tilde{U}_{j_{2}}^{2}=FFT(|iFFT(\text{periodize}(\tilde{U}_{j_{1}}^{2},j_{2}))|)$\;
           %$\tilde{U}_{j_{2}}^{2}=FFT(\tilde{U}_{j_{2}}^{2})$\;
          % $\tilde{U}_{j_{2}}^{2}=\hat{\tilde{\phi}}_{J}\odot\tilde{U}_{j_{2}}^{1}$\;
           $\tilde{U}_{j_2}^{2}=\hat{\tilde{\phi}}_{J}\odot\tilde{U}_{j_{2}}^{2}$\;
          $S_{J}^{2}x[\lambda_{1},\lambda_{2}]=iFFT(\text{periodize}(\tilde{U}_{j_2}^{2},J))$\;
        }
      }
      
    \caption{Pseudo-code of the algorithm used in PyScatWave.}\label{algo}
    \end{edit}
\end{algorithm}

To this end, we propose to perform an “infix traversal” of the tree of computations of the scattering. We introduce $\{\tilde{U}_{j}^{1},\tilde{U}_{j}^{2}\}_{j\leq J}$, which are two sequences of temporary  variables of length $\{\frac{N}{2^{j}}\}_{j\leq J}$ and a vector $\tilde{U}_{0}^{0}$ of length $N$. The total amount of memory that will be used is at most $5N^{2}$. Here, a color image of size $N=256$  corresponds to at most approximately 0.98M coefficients. It divides the memory usage by at least 2 and permits us to scale processing to ImageNet. Algorithm~\ref{algo} presents the algorithm we used in our implementation, dubbed PyScatWave. Table~\ref{speed} demonstrates the speed-up for different values of tensors on a TitanX, compared with ScatNetLight~\cite{oyallon2015deep}. 

We also note that in the case of training hybrid networks it is possible to store the computed scattering coefficients for a dataset via a cache. In this case, it is possible to obtain a speedup by a large factor since no extra computations are required to compute the earlier layers as optimization of the network proceeds. These early layers are often the most computationally expensive in comparison with deeper layers.

%\section{Analysis of the Scattering Transform}

\subsection{Reconstruction from the Scattering Coefficients}
\label{sec:reconst}
\addition
Reconstruction of an image from a scattering representation can be critical for permitting it's use in applications such as image generation. It also permits to obtain insights into the representation.
We describe a simple method to reconstruct an image from its order 2 scattering representation. Several works \cite{dokmanic2016inverse,bruna2013audio}  proposed to synthesize textures and stochastic processes from their expected scattering coefficients. In the case of stationary processes, the final local averaging of a scattering transform allows the building of an unbiased estimator of the expected scattering coefficients, and the smallest variance is achieved using the largest windows size of invariance, i.e.\ the full image. This does not hold in the case of natural images, which do not correspond to stationary processes, and thus, global invariance to translation is not desirable because it loses spatial localization information. We show a straightforward approach can yield competitive reconstruction.
 
\begin{figure}%[h]
\begin{center}
%\framebox[4.0in]{$\;$}
%\fbox{\rule[-.5cm]{0cm}{4cm}   
%\setlength\tabcolsep{1.5pt}
\begin{tabular}{cc}
\includegraphics[width=90px]{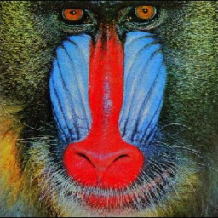}& \includegraphics[width=90px]{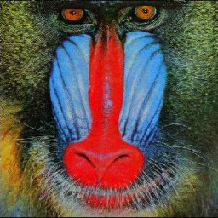}\\
\small Original image&\small  $2,7\times 10^{-3},6\times 10^{-2}$\\
%&$\text{err}(S_2)=7\times 10^{-3}$\\
%&$\text{err}(x)=6\times 10^{-2}$\\
\\

 \includegraphics[width=90px]{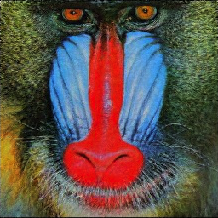}& \includegraphics[width=90px]{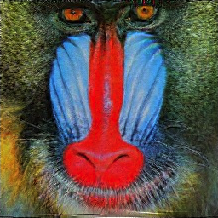}\\
$3,7\times 10^{-3},9\times 10^{-2}$&$4,7\times 10^{-3},1.1\times 10^{-1}$\\
%$J=3$& $J=4$\\
%$\text{err}(S_3)=7\times 10^{-3}$ &$\text{err}(S_4)=7\times 10^{-3}$\\
%$\text{err}(x)=9\times 10^{-2}$ &$\text{err}(x)=1.1\times 10^{-1}$ \\

\\

 \includegraphics[width=90px]{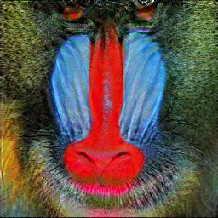}& \includegraphics[width=90px]{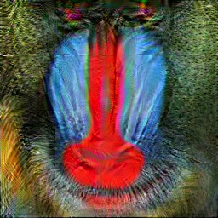}\\
$5,7\times 10^{-3},1.4\times 10^{-1}$&$6,7\times 10^{-3},1.7\times 10^{-1}$\\
%$J=5$& $J=6$\\
%$\text{err}(S_5)=7\times 10^{-3}$ &$\text{err}(S_6)=7\times 10^{-3}$\\
%$\text{err}(x)=1.4\times 10^{-1}$ &$\text{err}(x)=1.7\times 10^{-1}$ \\
\\
\includegraphics[width=90px]{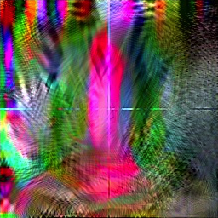}&\includegraphics[width=90px]{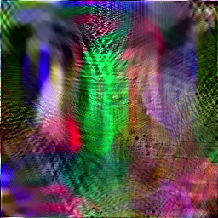}\\
$7,1.5\times 10^{-2},4.0\times 10^{-1}$&$8,1.0\times 10^{-2},4.4\times 10^{-1}$
% $J=7$&$J=8$\\
%$\text{err}(S_7)=1.5\times 10^{-2}$&$\text{err}(S_8)=1.0\times 10^{-2}$\\
%$\text{err}(x)=4.0\times 10^{-1}$&$\text{err}(x)=4.4\times 10^{-1}$\\
\end{tabular}
\caption{Reconstructed images with subcaption indicating $J,\text{err}(S_J),\text{err}(x)$. See Section~\ref{sec:reconst} for details of the reconstruction approach.}
\label{reco}
\end{center}
\end{figure}

\begin{edit} The method used \citep{bruna2013audio,bruna2013scattering} consists in minimizing the $\ell_2$ reconstruction error between an input image $x$ and a candidate $\hat x$:\end{edit}
\[\hat x = \arg\inf_y \Vert S_J x - S_J y\Vert_2\]
This is achieved via a gradient descent, without however any (known) theoretical guarantees of convergence to the original signal. Computations are made possible thanks to the auto-differentiation tool of PyTorch. In this setting, we chose the optimizer Adam. The initial image is initialized as a white noise with variance $10^{-4}$ and is represented in the YUV space because it decorrelates approximatively the color channels and the intensity channels, and we observed it leads to better reconstruction. The algorithm converges to a visually reasonable solution after $200$ iterations, the loss reaching a plateau, and there is no extra-regularization or parametrization because empirically this has not yielded better reconstruction. Results are displayed in  Figure~\ref{reco} for different values of $J$ and an image $x$ of size $256^2$. For each reconstruction, we evaluate its quality by computing the relative error of reconstruction with the original signal $\text{err}(x)$, and its distance in the scattering space $\text{err}(S_J)$,
\[\text{err}(x)=\frac{\Vert \hat x- x \Vert}{\Vert x \Vert} \quad \text{and} \quad \text{err}(S_J)=\frac{\Vert S_J\hat x- S_Jx \Vert}{\Vert S_Jx \Vert}.\]

We demonstrate good reconstruction in the case of $J=2,3,4$ and we show that numerically, by $J\geq5$, the obtained images are rather different from the original image due to the averaging loss. The attributes that are not well reconstructed are blurry and not at the appropriate spatial localization, which seems to indicate they have been lost by the spatial averaging. For $J<7$, the Scattering coefficients are almost identical, however, for $J=5,6$ several corners and borders of the images are not well recovered, which indicates it is possible to find very different images with similar scattering coefficients. An open question is to understand if cascading more wavelet transforms  could recover this information. For $J\geq 7$, the reconstructed signals are very different, only several textures seem to have been recovered and the color channels are decorrelated. Furthermore, the case $J=7$ exhibits strong artifacts from the large scale wavelet, which is linked to the implementation of the wavelet transform.

Due to this lack of localization and ability to discriminate, in the following sections we combine CNNs with a scattering transform with scales $J<5$, and therefore filters of width less than $2^5=32$ pixels.
\color{black}

%% file: supervision_on_top.tex
\subsection{Cascading a Supervised Architecture on Top of Scattering}
\label{sec:reconst}
We now motivate the use of a supervised architecture on top of a scattering network. 
%Cascading supervision on a predefined representation is necessary to obtain good accuracy on challenging benchmarks \cite{}. Indeed, 
Scattering transforms have yielded excellent numerical results \cite{bruna2013invariant} on datasets where the variabilities are completely known, such as MNIST or FERET. In these task, the problems encountered are linked to sample and geometric variance and handling these variances leads to solving these problems. However, in classification tasks on more complex image datasets, such variabilities are only partially known as there are also non geometrical intra-class variabilities. Although applying the scattering transform on datasets like CIFAR-10 or CalTech leads to nearly state-of-the-art results in comparison to other unsupervised representations, there is a large gap in performance when comparing to supervised representations \cite{oyallon2015deep}. CNNs fill in this gap. Thus we consider the use of deep neural networks utilizing generic scattering representations in order to learn more complex invariances than geometric ones alone.

%Furthermore neural networks have been shown to yield excellent performance even when used as classifiers \cite{}, when the features are complex. We thus consider the use of neural networks as classifiers utilizing generic scattering representations in order to reduce more complex variabilities than geometric ones.

%features that enrich the scattering transfrom allowing to learn additional variation from the dataset. 

 Recent works \citep{mallat2016understanding,bruna2013learning,jacobsen2017multiscale} have suggested that  deep networks could  build an approximation of the group of symmetries of a classification task and apply transformations along the orbits of this group, like convolutions. This group of symmetry corresponds to some of the non-informative intra class variabilities, which must be reduced by a supervised classifier. \citep{mallat2016understanding} motivates that each layer corresponds to an approximated Lie group of symmetry, and this approximation is progressive in the sense that the dimension of these groups is increasing with depth. For instance, the main linear Lie group of symmetry of an image is the translation group, $\mathbb{R}^2$. In the case of a wavelet transform obtained by rotation of a mother wavelet, it is possible to recover a new subgroup of symmetry after a modulus non-linearity, the rotation $SO_2$, and the group of symmetry at this layer is the roto-translation group: $\mathbb{R}^2 \ltimes SO_2$. If no non-linearity was applied, a convolution along $\mathbb{R}^2 \ltimes SO_2$ would be equivalent to a spatial convolution. Discovering explicitly the next new and non-geometrical groups of symmetry is however a difficult task \cite{jacobsen2017multiscale}; nonetheless, the roto-translation group seems to be a good initialization for the first layers. In this work, we investigate this hypothesis and avoid learning those well-known symmetries.

%\textcolor{red}{EB: This paragraph above might be too much fanciness without clearly adding to the point of why it's a good initilization. some reviewer might notice that}

Thus, we consider two types of cascaded deep networks on top of scattering. The first, referred to as the \textit{Shared Local Encoder} (SLE), learns a supervised local encoding of the scattering coefficients. We motivate and describe the SLE in the next subsection as an intermediate representation between unsupervised local pipelines, widely used in computer vision prior to 2012, and modern supervised deep feature learning approaches. 
%They permit building local descriptors with non-overlapping windows, for classification task. 
The second, referred to as a hybrid CNN, is a cascade of a scattering network and a standard CNN architecture, such as a ResNet \cite{he2015deep}. In the sequel we empirically analyse hybrid CNNs, which allow us to greatly reduce the spatial dimensions on which convolutions are learned and can reduce sample complexity.   
  
%In the next section we proposes a deep architecture, \textcolor{red}{that permits  building a bridge between unsupervised local pipelines, prior to 2012, and supervised architectures , posterior to 2012.} %that permits obtaining interpretability of its first layer, while relating post 2012 CNNs architecture, with prior 2012 handcrafted architectures.

%Notably, the first layer $F_1$  of this deep convolutional network is structured by its input, the scattering representation. The nature of this operator and the features selected by this supervised algorithm will be discussed in the section \ref{}.

%% file: supervised_encoding.tex
%This is consistent with the notion of symmetry, as it implies that one should only perform a parallel transport along the channels axis \ref{mallat}.
%\textcolor{red}{THIS NEED TO BE IMPROVED}

%EB: I tried to edit this part but its not coherent. most of the sentences are not needed to state something simple (CNN have big receptive field). I have revised it if you are happy you can delete this slash if

%In standard convolutional neural networks the representations at a specific spatial location depends upon large parts of the initial input image. Deep neural networks are a cascade of linear and point-wise non- linear operators. In CNNs, as translation is one of the main symmetries of the classification task, each operator is covariant with the action of translation. The spatial support of those convolutional  operators is often chosen small, with the standard size of 3\cite{vgg zisserman people}.

We now discuss the spatial support of different approaches, in order to motivate our local encoder for scattering. In CNNs constructed for large scale image recognition, the representations at a specific spatial location and depth depend upon large parts of the initial input image and thus mixes global  information. For example, in \cite{krizhevsky2012imagenet}, the effective spatial support of the corresponding filter is already 32 pixels (out of 224) at  depth 2. The specific representations derived from CNNs trained on large scale image recognition are often used as representations in other computer vision tasks or datasets \cite{yosinski2014transferable,zeiler2014visualizing}. %The mix of global and local information may not always be desirable as one can lose interpretablity of the encoded features at a spatial location. 

On the other hand prior to  2012 local encoding methods led to state of the art performance on large scale visual recognition tasks \cite{sanchez2011high}. In these approaches local neighborhoods of an image were encoded using method such as SIFT descriptors \cite{lowe1999object}, HOG \cite{dalal2005histograms}, and wavelet transforms \cite{serre2004realistic}. They were also often combined with an unsupervised encoding, such as sparse coding \cite{boureau2011ask} or Fisher Vectors (FVs) \cite{sanchez2011high}. Indeed, many works in classical image processing or  classification \cite{koenderink1999structure,boureau2011ask,sanchez2011high,perronnin2015fisher} suggest that local encodings of an image are efficient descriptions. Additionally for some algorithms that rely on local neighbourhoods, the use of local descriptors is essential \cite{lowe1999object}. Observe that a representation based on local non overlapping spatial neighborhood is simpler to analyze, as there is no ad-hoc mixing of spatial information. Nevertheless, in large scale classification, this approach was surpassed by fully supervised learned methods \cite{krizhevsky2012imagenet}.

We show that it is possible to apply a similarly local, yet supervised encoding algorithm to a scattering transform, as suggested in the conclusion of \cite{perronnin2015fisher}. First observe that at each spatial position $u$, a scattering coefficient $S(u)$ corresponds to a descriptor of a local neighborhood of spatial size $2^J$. As explained in the first Subsection \ref{scatnet}, each of our scattering coefficients are obtained using a stride of $2^J$, which means the final representation can be interpreted as a non-overlapping concatenation of  descriptors. Let $f$  be   a cascade of fully connected layers that we identically apply on each $Sx(u)$.  Then $f$ is a cascade of CNN  operators with spatial support size $1\times 1$, thus we write $fSx\triangleq \{f(Sx(u))\}_u$. In the sequel, we do not make any distinction between the $1\times1$ CNN operators and the operator acting on $Sx(u),\forall u$. We refer to $f$ as a \textit{Shared Local Encoder}. We note that similarly to $Sx$, $fSx$ corresponds to non-overlapping encoded descriptors. To learn a supervised classifier on a large scale image recognition task, we cascade fully connected layers on top of the SLE.

Combined with a scattering network, the supervised SLE, has several advantages. Since the input corresponds to scattering coefficients whose channels are structured, the first layer of $f$ is structured as well. We further explain and investigate this first layer in Subsection \ref{sec:sle_first_sec}. Unlike standard CNNs, there is no linear combination of spatial neighborhoods of the different feature maps, thus the analysis of this network need only focus on the channel axis.  Observe that if $f$ was fed with raw images, for example in gray scale, it could not build any non-trivial operation except separating different level sets of these images.  
%We note it is not possible to use an average pooling as it will remove any spatial localization information for the supervised task.

In the next section, we investigate empirically this supervised SLE trained on the ILSVRC2012 dataset.

%% file: SLE.tex
\begin{figure}
\begin{center}
\begin{tikzpicture}

\node at (0,-0.6) (d1) { $\vdots$};
\node at (0,1.8)  (d21) { $\vdots$};
\node at (3.5,-0.6) (d12) { $\vdots$};
\node at (3.5,1.8)  (d22) { $\vdots$};
\node at (0,1.2) [minimum width=0.5cm,line width=0] (S1) {\footnotesize $Sx(u-2^J)$};
\node at (0,0.6) [minimum width=0.5cm,line width=0] (S2) {\footnotesize  $Sx(u)$};
\node at (0,0) [minimum width=0.5cm,line width=0] (S3) {\footnotesize $Sx(u+2^J)$};

\node at (4.5,0.6) [minimum width=0.5cm,line width=0,draw] (F_4) {$F_4$};
\node at (5.5,0.6) [minimum width=0.5cm,line width=0,draw] (F_5) {$F_5$};
\node at (6.5,0.6) [minimum width=0.5cm,line width=0,draw] (F_6) {$F_6$};
\node at (7.2,0.6)  (F_7) {};
\foreach \i in {0,...,2}
{

\foreach \j in {1,...,3}
{

\node at (1*\j+0.5,0.6*\i) [minimum width=0.5cm,line width=0,draw] (\j_F_\i) {$F_\j $};
%\node at (-0.3,0.6) [minimum width=1cm,line width=0] (x) {$Sx(u)$};
%\node at (0,0) [minimum width=1cm,line width=0] (x) {$Sx(u+1)$};
}
}

\foreach \i in {0,...,2}
{

\foreach \j[evaluate = \j as \jp using int(\j+1)] in {1,...,2} 
{
\draw[->,black] (\j_F_\i) -- (\jp_F_\i);

}
}

\foreach \i in {0,...,2}
{
\draw[->,black] (3_F_\i) -- (F_4);
}

\draw[->,black] (S1) -- (1_F_2);
\draw[->,black] (S2) -- (1_F_1);
\draw[->,black] (S3) -- (1_F_0);

\draw[->,black,dotted] (d22.east) -- (F_4);
\draw[->,black,dotted] (d12.east) -- (F_4);

\draw[->,black] (F_4) -- (F_5);
\draw[->,black] (F_5) -- (F_6);
\draw[->,black] (F_6) -- (F_7);
%\node at (0,0.7) [minimum width=1cm,draw,align=center] (s2) {\small $F_1$};
%\node at (0,1.4) [minimum width=1cm,draw,align=center] (W2) {\small $F_2$};
\end{tikzpicture}
%\fbox{\rule{0pt}{2in} \rule{.9\linewidth}{0pt}}
\end{center}
   \caption{Architecture of the SLE, which is a cascade of 3 $1\times 1$ convolutions followed by 3 fully connected layers. The ReLU non-linearities are included inside the $F_i$ blocks.}
\label{fig:model}
\end{figure}

\begin{table}
\begin{center}
\begin{tabular}{|l|c|c|c|}
\hline
\bf Method  &\bf Top 1 &\bf Top 5  \\
\hline
%\small \bf   Scattering hybrids &  & \\

%\small \bf  Unsupervised   &  &  \\

FV + FC    \cite{perronnin2015fisher} &55.6  & 78.4 \\% &44.4  & 21.6 \\
FV + SVM   \cite{sanchez2011high}          & 54.3 & 74.3\\%45.7 & 25.7 \\
%\hline
%\small \bf  Supervised  and Hybrids  & &  \\
AlexNet  & 56.9 &\bf 80.1\\%43.1 & \bf 19.9\\
Scat + SLE  & \bf 57.0&79.6\\%\bf 43.0 &20.4 \\
%GoogleNet& 32.8&11.0\\
\hline
\end{tabular}
\end{center}
\caption{Top 1 and Top 5 percentage accuracy reported from one single crop on ILSVRC2012. We compare to other local encoding methods, and the Shared Local Encode (SLE) outperforms them (see Sec.~\ref{sec:sle_exp} for experiment details).  \cite{perronnin2015fisher} single-crop result was provided by private communication.}\label{res_1x1}
\end{table}

We first describe our training pipeline, which is similar to \cite{zagoruyko2016wide}. We trained our network for 90 epochs to minimize the standard cross entropy loss, using SGD with momentum 0.9 and a batch size of 256. We used a weight decay of $1\times10^{-4}$. The initial learning rate is $0.1$, and is decreased by a factor of 10 at epochs $30$, $50$, $70$, and $80$. During the training process, each image is randomly rescaled, cropped, and flipped as in \citep{he2015deep}. The final crop size is $224\times 224$. At testing, we rescale the image to a size of $256 \times 256$, and extract a center crop of size $224\times 224$. 

We use an architecture which consists of a cascade of a scattering network, a SLE $f$, followed by fully connected layers. Figure \ref{fig:model} describes our architecture. We select the parameter $J=4$ for our scattering network, which means the output representation has size $\frac{224}{2^4}\times\frac{224}{2^4}=14\times 14$ spatially and 1251  channels. $f$ is implemented as 3 layers of 1x1 convolutions $F_1,F_2,F_3$ with layer size 1024. There are 2 fully connected layers of ouput size 1524. For all learned layers we use batch normalization \cite{ioffe2015batch} followed by a ReLU \cite{krizhevsky2012imagenet} non-linearity. We compute the mean and variance of the scattering coefficients on the whole of ImageNet, and standardized each spatial scattering coefficients with them. %Since we observed that the Scattering features can suffer from a bad conditionning, we also apply a batch normalization at the ouput of the Scattering network, with a  momentum equal to 0.9.

%pared to other analogous models, several of them corresponding to state-of-the-art approaches in 2012 and a 2015 one\cite{perronnin2015fisher}.
Table \ref{res_1x1} reports our numerical accuracies obtained with a single crop at testing, compared with local encoding methods, and AlexNet, which was the state-of-the-art approach in 2012. We obtain 20.4\% at Top 5 and 43.0\% Top 1 errors. The performance is analogous to the AlexNet \cite{krizhevsky2012imagenet}. In term of architecture, our hybrid model is analogous, and comparable to that of \cite{sanchez2011high,perronnin2015fisher}, for which SIFT features are extracted followed by FV \cite{sanchez2013image}  encoding. Observe the FV is an unsupervised encoding compared to our supervised encoding. Two approaches are then used: the  spatial localization is handled either by a Spatial Pyramid Pooling \cite{lazebnik2006beyond}, which is then fed to a linear SVM, or the spatial variables are directly encoded in the FVs and classified with a stack of four fully connected layers. This last method is a major difference with ours, as the obtained descriptor does not have a spatial indexing  anymore which are instead quantized. Furthermore, in both case, the SIFT are densely extracted which correspond to approximatively $2\times 10^4$ descriptors, whereas in our case, only $14^2=196$ scattering coefficients are extracted. Indeed, we tackle the non-linear aliasing (due to the fact that the scattering transform is not oversampled) via random cropping during training, enabling invariance to small translations. In Top 1, \cite{sanchez2011high} and \cite{perronnin2015fisher} obtain error rates of 44.4\% and 45.7\%, respectively. Our method brings a substantial improvement of 1.4\% and 2.7\%, respectively.

The BVLC AlexNet\footnote{https://github.com/BVLC/caffe/wiki/Models-accuracy-on-ImageNet-2012-val} obtains a  of 43.1\%  single-crop Top 1  error, which is nearly equivalent to the 43.0\% of our SLE network. The AlexNet has 8 learned layers and as explained before, large receptive fields. On the contrary, our training pipeline consists in 6 learned layers with constant receptive field of size $16 \times 16$, except for the fully connected layers that build a  representation mixing spatial information from different locations. This is a surprising result, as it seems to suggest contextual information is only necessary at the very last layers, to reach AlexNet accuracy.

We study briefly the local SLE, which only has a spatial extent of $16 \times 16$, as a generic local image descriptor. We use the Caltech-101 benchmark which is a dataset of 9144 images and 102 classes. We followed the standard protocol for evaluation \cite{boureau2011ask} with 10 folds and evaluate per class accuracy with 30 training samples per class, using a linear SVM used with the SLE descriptors. Applying our raw scattering network leads to an accuracy of $62.8\pm0.7$, and the output features from $F_1$, $F_2$, and $F_3$ bring an absolute improvement of $13.7$, $17.3$, and $20.1$, respectively. The accuracy of the final SLE descriptor is thus  $82.9\pm0.4$, similar to that reported for the AlexNet final layer in \cite{zeiler2014visualizing} and sparse coding with SIFT \cite{boureau2011ask}. However in both cases spatial variability is removed, either by Spatial Pyramid Pooling \cite{lazebnik2006beyond}, or the cascade of large filters. By contrast, the concatenation of SLE descriptors are completely local.  Similarly, the scattering network combined with ResNet-10 introduced in the next section, and followed by a linear SVM achieves 87.7 on Caltech-101, yet this descriptor is not local.

%% file: interpretating.tex
\begin{figure}
\begin{center}
\includegraphics[width=\linewidth]{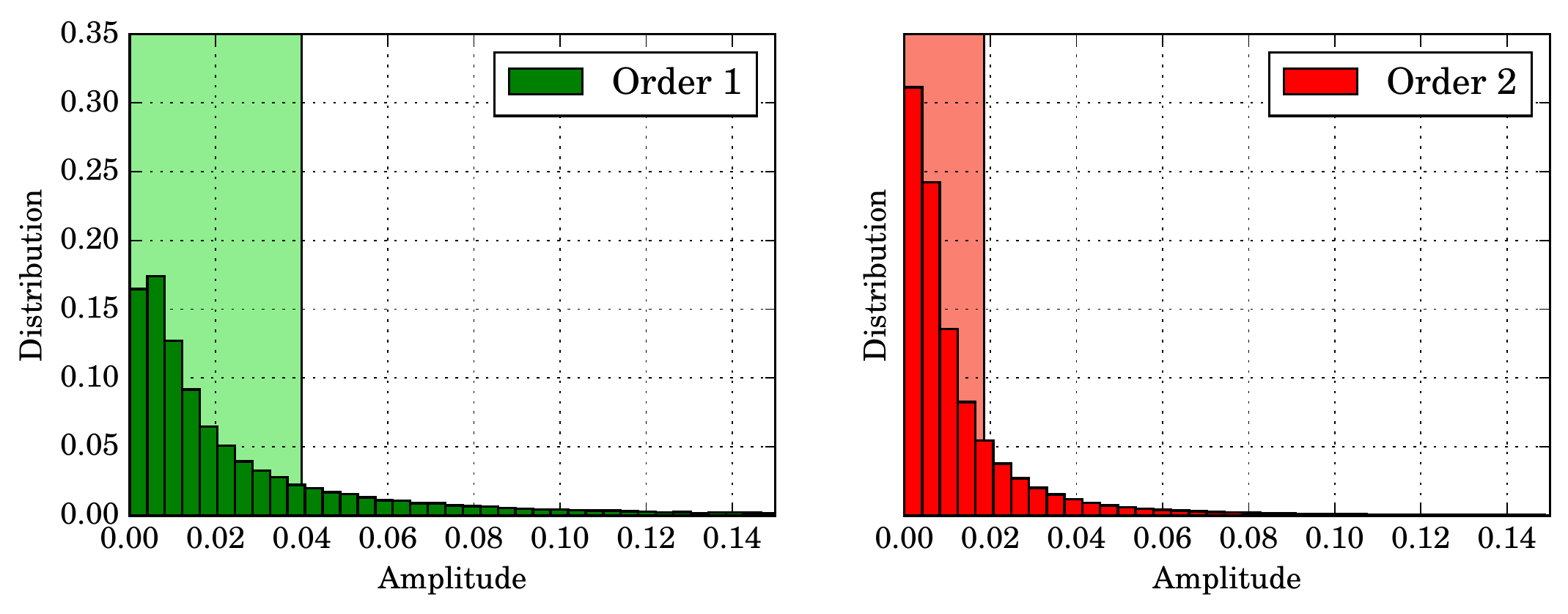} 
%\fbox{\rule{0pt}{2in} \rule{.9\linewidth}{0pt}}
\end{center}
   \caption{Histogram of  $\hat F_1$ amplitude for first and second order coefficients. The vertical lines indicate a threshold that is used in Subsection \ref{interpretation} to sparsify $\hat F_1$.}
\label{fig:histo}
\end{figure}

%\begin{figure}
%\begin{center}
%\includegraphics[width=.8\linewidth]{little_wood.eps} 
%\end{center}
%   \caption{Propagated $\Omega_2\{F\}$ from Eq. \ref{eq1} for given angular frequencies  $(\omega_{\theta_1},%\omega_{\theta_2})$.}
%\label{fig:littlehood2}
%\end{figure}

%Many works have studied the Alexnet architectures . The simplicity of our model and the fact that it is built upon allow us to show insights on the nature of the operations required to obtain strong classification performance. 

Finding structure in the kernel of the layers of depth less than $2$ \cite{waldspurger2015these,zeiler2014visualizing} is  a complex task, and few empirical analyses exist that shed light on the structure \cite{jacobsen2017multiscale} of deeper layers. A scattering transform with scale $J$ can be interpreted as a CNN with depth $J$ \cite{oyallon2015deep}, whose  channels indexes correspond to different scattering frequency indexes, which is a structuration. This structure is consequently inherited by the first layer $F_1$ of our SLE $f$. We analyse $F_1$ and show that it explicitly builds invariance to local rotations, and also that the Fourier bases associated to rotations are a natural bases of our operator. It is a promising direction to understand the nature of the next two layers. %Thus to a greater extent than the analogously performing AlexNet the behavior of our SLE network can be understood \textcolor{red}{I wrote it but is it too strong this statement? I like it if not}.

We first establish some mathematical notions linked to the rotation group that we use in our analysis. For the sake of clarity, we do not consider the roto-translation group.
For a given input image $x$, let $r_\theta . x(u)\triangleq x(r_{-\theta }(u))$ be the image  rotated by angle $\theta$, which corresponds to  the linear action of rotation on images. Observe the scattering representation is  covariant with the rotation in the following sense:
\begin{align*}
S^1(r_\theta.x)(\theta_1,u) &= S^ 1x(\theta_1-\theta,r_{-\theta}u)\triangleq r_\theta.(S^1x)(\theta_1,u) ,\\
S^2(r_\theta.x)(\theta_1,\theta_2,u)&=S^2x(\theta_1-\theta,\theta_2-\theta,r_{-\theta}u) \\
&\triangleq r_\theta.(S^2x)(\theta_1,\theta_2,u) .
\end{align*}
Additionally, in the case of the second order coefficients, $(\theta_1,\theta_2)$ is covariant with rotations, but $\theta_2-\theta_1$ is an invariant to rotation that corresponds to a relative rotation. 

%The rotations are a compact group, thus it is possible \cite{complex haarmonic anlasysis book like watman} to define a Fourier transform along rotation, and even  a scattering transform \cite{laurent sifre} along the more complex roto-translation group \textcolor{red}{This sentence has grammatical error, but E doesnt know enough of this math to fix it}.
The unitary representation framework \cite{sugiura1990unitary} permits the building of a Fourier transform on a compact group, such as rotations. It is even possible to build a scattering transform on the roto-translation group \cite{sifre2013rotation}. Fourier analysis permits the measurement of the smoothness of the operator and, in the case of a CNN operator, it is a natural basis.%: for a function $f:\theta \rightarrow f(\theta)$, we write $\hat f: \omega_\theta \rightarrow \hat f(\omega_\theta)$ its Fourier transform.
%Observe that applying a rotation $r_\theta . x=x(r_{-\theta }u)$ translates the coefficients, ie:

%However, for $S_2x$, the natural set of coordinates that gives a rotational invariant for angles is in fact given by:
%\[\tilde S_2 x(u,\theta_1,\alpha)=S_2x(u,\theta_1,\theta_1+\alpha)\]
%where $\alpha$ corresponds to a relative angle, which naturally leads to:
%\[\tilde S_2(r_\theta.x)(u,\theta_1,\alpha)=\tilde S_2x(r_{-\theta}u, \theta_1-\theta,\alpha)\triangleq r_\theta.(S_2x)(u,\theta_1,\alpha)\]

We can now numerically analyse the nature of the operations performed along angle variables by the first layer $F_1$ of $f$, with output size $K=1024$. Let us define as $\{F^0_1S^0x,F^1_1S^1x,F^2_1S^2x\}$ the restrictions of $F_1$ to the order 0, 1, and 2 scattering coefficients respectively. Let $1\leq k \leq K$ be an index of a feature channel and $1\leq c\leq 3$ be the color index. In this case, $F^0_1S^0x$ is simply the weights associated to the smoothing $S_0x$. $F^1_1S^1x$ depends only on $(k,c,j_1,\theta_1)$, and $F^2_1$ depends on $(k,c,j_1,j_2,\theta_1,\theta_2)$. We would like to characterize the smoothness of these operators with respect to the variables $(\theta_1,\theta_2)$, because $Sx$ is covariant to rotations.

To this end, we define by $\hat F_1^1$, $\hat F_1^2$ the Fourier transform of these operators along the variables $\theta_1$ and $(\theta_1,\theta_2)$ respectively. These operator are expressed in the tensorial frequency domain, which corresponds to a change of basis. In this experiment, we normalized each filter of $F$ such that they have a $\ell_2$ norm equal to 1, and each order of the scattering coefficients are normalized as well. Figure \ref{fig:histo} shows the distribution of the amplitude of $\hat F_1^1,\hat F_2^2$. We observe that the distribution is shaped as a Laplace distribution, which is an indicator of sparsity. 

To illustrate that this is a natural basis we explicitly sparsify this operator in its frequency basis and verify that empirically the network accuracy is minimally changed. We do this by thresholding  by $\epsilon$ the coefficients of the operators in the Fourier domain. Specifically we replace the operators $\hat F_1^1$, $\hat F_1^2$ by $1_{|\hat F_1^1|>\epsilon}\hat F_1^1$ and $1_{|\hat F_1^2|>\epsilon}\hat F_1^2$. We select an $\epsilon$ that sets $80\%$ of the coefficients to 0, which is illustrated in Figure \ref{fig:histo}. \textit{Without retraining} our network performance degrades by only an absolute value of $2\%$ worse on Top 1 and Top 5 ILSVRC2012. We have thus shown that this basis permits a sparse approximation of the first layer, $F_1$. We now show evidence that this operator builds an explicit invariant to local rotations.

\begin{figure}
\begin{center}
\includegraphics[width=\linewidth]{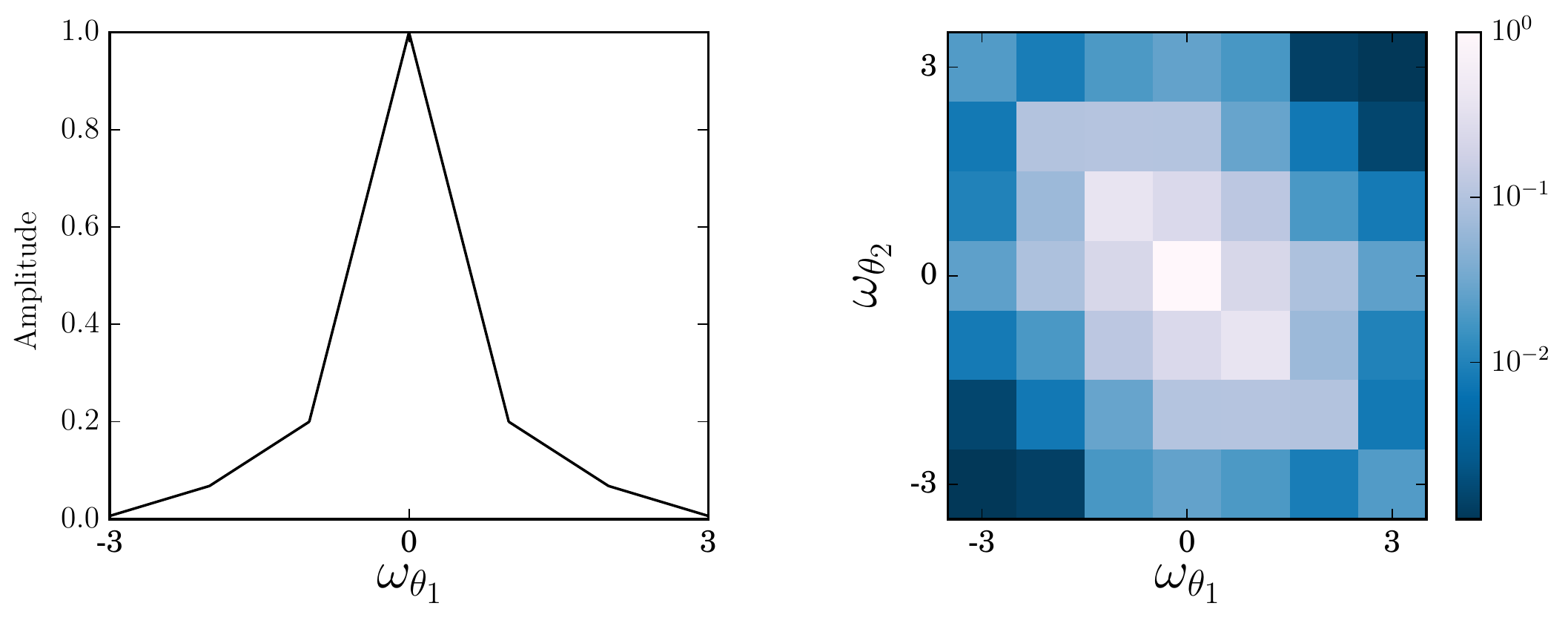} 
\end{center}
   \caption{Energy $\Omega_1\{F\}$  (left)  and $\Omega_2\{F\}$ (right) from Eq. \ref{eq1} for given angular frequencies.}
\label{fig:littlehood1}
\end{figure}

To aid our analysis we introduce the following quantities:
\setlength{\belowdisplayskip}{0pt} \setlength{\belowdisplayshortskip}{0pt}
\setlength{\abovedisplayskip}{0pt} \setlength{\abovedisplayshortskip}{0pt}
%\begin{align}
\begin{equation}
\Omega_1\{F\}(\omega_1)\triangleq\sum_{k,j_1,c}|\hat{F}^1_1(k,c,j_1,\omega_{\theta_1})|^2 ,
\label{eq1}
\end{equation}
%\begin{align}
\begin{equation*}
\Omega_2\{F\}(\omega_{\theta_1},\omega_{\theta_2})\triangleq\sum_{k,c,j_1,j_2}|\hat{F}^2_1(k,c,j_1,j_2,\omega_{\theta_1},\omega_{\theta_2})|^2 .
%\label{eq2}
\end{equation*}
%\end{align}
They correspond to the energy propagated by $F_1$ for a given frequency, and quantify the smoothness of our first layer operator w.r.t.\ the angular variables. Figure \ref{fig:littlehood1} shows variation of  $\Omega_1\{F\}$ and $\Omega_2\{F\}$ as a function of the frequencies. For example, if $F_1^1$ and $F_1^2$ were convolutional along $\theta_1$ and $(\theta_1,\theta_2)$, these quantities would correspond to their respective singular values. One sees that the energy is concentrated in the low frequency domain, which indicates that $F_1$ builds explicitly an invariant to local rotations.

%% file: small.tex
%Supervised deep networks trained on small datasets can easily overfit, for instance in the case of medical imaging where little data are available. Semisupervised algorithms exhibit good performances \citep{salimans2016improved}, but it requires a large amount of unlabeled data to work. In this subsection, we show the benefit of using scattering in a framework where those data are not available. We demonstrate that scattering does prevent overfitting on small datasets, while keeping the same architecture and training methodology: this saves time to design an architecture.
%A major application of a hybrid representation is in the setting of limited data. Here the learning algorithm is limited in the variations it can observe or learn from the data, such that introducing invariances without a loss of discriminative performance can substantially improve performance.
%In this subsection, we show the benefit of using scattering in a framework where those data are not available. We demonstrate that scattering does prevent overfitting on small datasets, while keeping the same architecture and training methodology: this saves time to design an architecture.

% We report as well state-of-the-arts results in the case of unsupervised learning and small data.
\subsection{Deep Hybrid CNNs on ILSVRC2012}
\label{hybimnet}
\begin{table}
\begin{center}
\begin{tabular}{|l|c|c|c|c|}
\hline
\bf Method  &\bf Top 1 &\bf Top 5 &\bf Params \\
\hline
AlexNet & 56.9&80.1&61M\\%43.1 & 19.9 & 61M\\
VGG-16 \cite{han2015learning}&68.5&88.7&138M\\%31.5 &11.3 & 138M\\
Scat + Resnet-10 (ours)&68.7&88.6&12.8M\\%31.3 & 11.4 & 12.8M \\
Resnet-18 & 68.9&88.8&11.7M\\%31.1  & 11.2 & 11.7M  \\
Resnet-200 \cite{zagoruyko2016wide} & \textbf{78.3} & \textbf{94.2} & 64.7M  \\%\textbf{21.7} & \textbf{5.8} & 64.7M  \\
\hline
\end{tabular}
\end{center}
\caption{ILSVRC-2012 validation accuracy (single crop) of hybrid scattering and 10 layer ResNet, a comparable 18 layer ResNet, and other well known benchmarks. We obtain comparable performance using a similar number of parameters while learning parameters at a spatial resolution of 28 $\times$ 28}
\label{tab:imagenet_full}
\end{table}

\begin{table}
\begin{center}
\begin{tabular}{|l|c|c|}
\hline
\bf Method & \bf Accuracy \\
  \hline 

\small \bf Unsupervised Representations   &  \\
\begin{edit}CKN \cite{mairal2014convolutional}\end{edit} & 82.2\\
Roto-Scat + SVM   \cite{oyallon2015deep}  &82.3\\
ExemplarCNN \cite{dosovitskiy2014discriminative} & 84.3 \\
DCGAN \cite{radford2015unsupervised}& 82.8 \\
Scat + FC   (ours)     & \bf 84.7\\
\hline
\small \bf Supervised and Hybrid    &  \\
Scat + WRN  (ours) & 93.1       \\
Highway  network \cite{srivastava2015highway}& 92.4 \\
All-CNN \cite{springenberg2014striving}& 92.8 \\
WRN 16 - 8 \cite{zagoruyko2016wide}  & 95.7\\
WRN 28 - 10 \cite{zagoruyko2016wide} &  \textbf{96.0} \\
\hline
\end{tabular}
\end{center}
\caption{Accuracy of scattering compared to similar architectures on CIFAR10. We set a new state-of-the-art in the unsupervised case and obtain competitive performance with hybrid CNNs in the supervised case.}\label{tab:CIFAR_Main}
\end{table}

We showed in the previous section that a SLE followed by FC layers can produce results comparable to AlexNet \cite{krizhevsky2012imagenet} on the ImageNet classification task. Here we consider cascading the scattering transform with a modern CNN architecture, such as ResNet \cite{zagoruyko2016wide,he2015deep}. We take ResNet-18 \cite{zagoruyko2016wide} as a reference and construct a similar architecture with only 10 layers on top of the scattering network.  We utilize a scattering transform with $J=3$ such that the CNN is learned over a spatial dimension of $28 \times 28$ and a channel dimension of 651 (3 color channels of 217 each). ResNet-18 typically has 4 residual stages of 2 blocks each which gradually decrease the spatial resolution \cite{zagoruyko2016wide}. Since we utilize the scattering as a first stage we remove two blocks from our model. The network is described in Table \ref{table:arch_imagenet}.

\newcommand{\blocka}[2]{
  \(\left[
      \begin{array}{c}
        \text{3$\times$3, #1}\\[-.1em]
        \text{3$\times$3, #1}
      \end{array}
    \right]\)$\times$#2
}
\newcommand{\blockb}[2]{
 \(\left[
      \begin{array}{c}
        \text{#1}\\[-.1em]
        \text{#1}
      \end{array}
    \right]\)$\times$#2
}
\newcommand{\convsize}[1]{#1$\times$#1}
\newcommand{\convname}[1]{#1}
\def\cellheight{0.34cm}

\begin{table}
  \centering
  \begin{tabular}{|c|c|c|}
    \hline
    Stage & Output size  & Stage details  \\
  %  \Xhline{2\arrayrulewidth}
    \hline
    scattering & \convsize{28}&   $J=3, 651$ channels \\
    \convname{conv1} & \convsize{28} & [256] \\[\cellheight]
    \convname{conv2} & \convsize{28} & \blockb{256}{2}\\[\cellheight]
    \convname{conv3} & \convsize{14} & \blockb{512}{2} \\
    avg-pool & $1\times1$ & [$14\times14$]  \\
    \hline
  \end{tabular}
  \vspace{0.2cm}
  \caption{Structure of Scattering and ResNet-10 architectures used in ImageNet experiments. Taking the convention of \cite{zagoruyko2016wide} we describe the convolution size and channels in the stage details.} 
  \label{table:arch_imagenet}
  % \vspace{-0.2cm}
\end{table}

\begin{table}
  \centering
  \begin{tabular}{|c|c|c|}
    \hline
    Stage & Output size & Stage details  \\
  %  \Xhline{2\arrayrulewidth}
    \hline
    scattering & $8\times8$,~$24\times24$ &   $J=2$ \\
    \convname{conv1} & \convsize{8},~\convsize{24} & 16$\times$k , 32$\times$k  \\[\cellheight]
    %\convname{conv2} & \convsize{8},$24\times24$  &Ablated,\blockb{16$\times$k}{2} \\[\cellheight]
    \convname{conv2} & \convsize{8},~\convsize{24}  &\blockb{32$\times$k}{$n$}\\[\cellheight]
    \convname{conv3} & \convsize{8},~\convsize{12}  &\blockb{64$\times$k}{$n$} \\
    avg-pool & $1\times1$ & [$8\times8$],~[$12\times12$]  \\
    \hline
  \end{tabular}
  \vspace{0.2cm}
  \caption{Structure of Scattering and Wide ResNet hybrid architectures used in small sample experiments. Network width is determined by factor $k$. For sizes and stage details if settings vary, we list CIFAR-10 and then the STL-10 network information. All convolutions are of size $3\times3$ and the channel width is shown in brackets for both the network applied to STL-10 and CIFAR-10. For CIFAR-10 we use $n=2$ and for the larger STL-10 we use $n=4$.}
  \label{table:arch_CIFAR}
  % \vspace{-0.2cm}
\end{table}

We use the same optimization and data augmentation procedure described in Section ~\ref{sec:sle_exp} but with decreases in the learning rate at 30, 60, and 80 epochs. We find that when both methods are trained with the same settings of optimization and data augmentation, and when the number of parameters is similar (12.8M versus 11.7 M) the scattering network combined with a ResNet can achieve analogous performance (11.4$\%$ Top 5 for our model versus 11.1$\%$), while utilizing fewer layers compared to a pure ResNet architecture. The accuracy is reported in Table \ref{tab:imagenet_full} and compared to other modern CNNs. 

This demonstrates both that the scattering networks does not lose discriminative power and that it can be used to replace early layers of standard CNNs. We also note that learned convolutions occur over a drastically reduced spatial resolution without resorting to pre-trained early layers, which can potentially lose discriminative information or become too task specific. 

%\textcolor{red}{We also note that training procedures for learning directly from images, including data augmentation, have been heavily optimized for networks trained directly on natural images.} 
\subsection{Deep Hybrid CNNs on CIFAR-10}
\label{sec:cifar}
We now consider the popular CIFAR-10 dataset consisting of color images composed of $5\times10^4$ images for training, and $1\times10^4$ images for testing divided into 10 classes. We use a hybrid CNN architecture with a ResNet built on top of the  scattering transform. 

For the scattering transform we used $J=2$ which means the output of the scattering stage will be $8\times8$ spatially and 243 in the channel dimension.  We follow the training procedure prescribed in \cite{zagoruyko2016wide} utilizing SGD with momentum of 0.9, batch size of 128, weigh decay of $5\times10^{-4}$, and modest data augmentation by using random cropping and flipping. The initial learning rate is 0.1, and we reduce it by a factor of 5 at epochs 60, 120 and 160. The models are trained for 200 epochs in total. We used the same optimization and data augmentation pipeline for training and evaluation in both case. We utilize batch normalization techniques at all layers which lead to a better conditioning of the optimization \citep{ioffe2015batch}. Table \ref{tab:CIFAR_Main} reports the accuracy in the unsupervised and supervised settings and compares them to other approaches.  

We compare to state-of-the-art approaches on CIFAR-10, all based on end-to-end learned CNNs. We use a similar hybrid architecture to the successful wide residual network (WRN) \cite{zagoruyko2016wide}. Specifically we modify the WRN of 16 layers, which consists of 4 convolutional stages. With $k$ denoting the widening factor, after the scattering output we use a first stage of $32\times k$. We add intermediate $1\times 1$ convolutions to increase the effective depth, without substantially increasing the number of parameters. Finally we apply a dropout of 0.2 as specified in \cite{zagoruyko2016wide}.
Using a width of 32 we achieve an accuracy of $93.1\%$. This is superior to several benchmarks but performs worse than the original ResNet \cite{he2015deep} and the wide ResNet \cite{zagoruyko2016wide}. We note that training procedures for learning directly from images, including data augmentation and optimization settings, have been heavily optimized for networks trained directly on natural images, while we use them largely out of the box. %We do believe there are regularization techniques, normalization techniques, and data augmentation techniques which can be designed specifically for the scattering networks.\todo{MB: should we say "This is an interesting area for future research."?}

\subsection{Limited samples setting}
\label{verysmall}
A major application of a hybrid representation is in the setting of limited data. Here the learning algorithm is limited in the variations it can observe or learn from the data, such that introducing a geometric prior can substantially improve performance. We evaluate our algorithm on the limited sample setting using a subset of CIFAR-10 and the STL-10 dataset.

\subsubsection{CIFAR-10}
We take subsets of decreasing size of the CIFAR dataset and train both baseline CNNs and counterparts that utilize the scattering as a first stage. We perform experiments using subsets of 1000, 500, and 100 samples, which are split uniformly amongst the 10 classes. 

We use as a  baseline the Wide ResNet \cite{zagoruyko2016wide}  of depth 16 and width 8, which shows near state-of-the-art performance on the full CIFAR-10 task in the supervised setting. This network consists of 4 stages of progressively decreasing spatial resolution detailed in  \cite[Table 1]{zagoruyko2016wide}. We construct a comparable hybrid architecture that removes a single stage and all strides, as the scattering already down-sampled the spatial resolution. This architecture is described in Table \ref{table:arch_CIFAR}.  Unlike the baseline, referred from here-on as WRN 16-8, our architecture has 12 layers and equivalent width, while keeping the spatial resolution constant through all stages prior to the final average pooling. We also incorporate the numerical results obtained via a VGG of depth 16  \cite{zagoruyko201592} for the sake of comparison.

We use the same training settings for our baseline, WRN 16-8, and our hybrid scattering and WRN-12. The settings are the same as  those described for CIFAR-10 in the previous section, with the only difference being that we apply a multiplier to the learning rate schedule and to the maximum number of epochs. The multiplier is set to 10, 20, and 100 for the 1000, 500, and 100 sample cases, respectively. For example the default schedule of 60, 120, and 160 epochs becomes 600, 1200, and 1600 for the case of 1000 samples and a multiplier of 10. Finally in the case of 100 samples we use a batch size of 32 in lieu of 128.

Table \ref{small} corresponds to the averaged accuracy over 5 different subsets, with the corresponding standard error. 
In this small sample setting, a hybrid network outperforms the purely CNN based baselines, particularly when the sample size is smaller. This is not surprising as we incorporate a geometric prior in the representation. 

%Nonetheless, a GAN performs better than a translation scattering with 3 fully connected layers; its performances is almost constant equal to $80\%$, which shows that this algorithm can adapt itself to the bias of the dataset. %For 1000 samples, there is a difference of 4.3$\%$ but only a difference of $2.9\%$ with 8000 samples.

\begin{table}
\begin{center}

\begin{tabular}{|l|c|c|c|c|}
\hline
\bf Method & \bf 100 & \bf 500 & \bf 1000 & \bf Full \\
  \hline 

WRN 16-8    & 34.7 $\pm$ 0.8  &    46.5 $\pm$1.4& 60.0 $\pm$1.8  & \bf 95.7\\
VGG 16 \cite{zagoruyko201592} & 25.5 $\pm 2.7$ & $46.2 \pm 2.6$ & $56 \pm 1.0$ & 92.6 \\ 
Scat + WRN  & \bf 38.9 $\pm$ 1.2 &\bf 54.7$\pm$0.6 &\bf 62.0$\pm\bf 1.1$& 93.1\\

\hline
\end{tabular}
\end{center}
\caption{Mean accuracy of a hybrid scattering in a limited sample situation on CIFAR-10 dataset. We find that including a scattering network is significantly better in the smaller sample regime of 500 and 100 samples.}
\label{small}
\end{table}

\subsubsection{STL-10}
\begin{table}
\begin{center}

\begin{tabular}{|l|c|c|}
\hline
\bf Method & \bf Accuracy \\
\hline

$\substack{\text{\small \bf  Supervised methods}}  $     &  \\
Scat + WRN 20-8 & \bf 76.0 $\pm$ 0.6\\
CNN\cite{swersky2013multi} &  70.1 $\pm$ 0.6\\
\hline
$\substack{\text{\small \bf  Unsupervised methods}}  $     &  \\
Exemplar CNN \cite{dosovitskiy2014discriminative}&  75.4 $\pm$ 0.3\\
%Deep Unsupervised \cite{hoffer2016deep}& 81.3$\pm$0.1\\
Stacked what-where AE \cite{StackedYann} & 74.33 \\
Hierarchical Matching Pursuit (HMP) \cite{bo2013unsupervised}& 64.5$\pm$1\\
Convolutional K-means Network \cite{coates2011selecting} & 60.1$\pm$1\\
\hline
\end{tabular}
\end{center}
\caption{Mean accuracy of a hybrid CNN on the STL-10 dataset. We  find that our model is  better in all cases even compared to those utilizing the large unsupervised part of the dataset.}

\label{tab:small_STL}
\end{table}
The STL-10 dataset consists of color images of size $96\times 96$, with only 5000 labeled images in the training set divided equally in 10 classes and 8000 images in the test set. The larger size of the images and the small number of available samples make this a challenging image classification task. The dataset also provides 100,000 unlabeled images for unsupervised learning. We do not utilize these images in our experiments, yet we find we are able to outperform all methods which learn unsupervised representations using these unlabeled images, obtaining very competitive results on the STL-10 dataset. 

We apply a hybrid convolutional architecture, similar to the one applied in the small sample CIFAR task, adapted to the size of $96\times 96$. The architecture is described in Table \ref{table:arch_CIFAR} and is similar to that used in the CIFAR small sample task. We use the same data augmentation as with the CIFAR datasets. We apply SGD with learning rate 0.1 and learning rate decay of 0.2 applied at epochs 1500, 2000, 3000, 4000. Training is run for 5000 epochs. We use at training and evaluation the predefined 10 folds of 1000 training images each, as given in \cite{StackedYann}. The averaged result is reported in Table \ref{tab:small_STL}. Unlike other approaches, we do not use the 4000 remaining training images to perform hyper-parameter tuning on each fold, as this is not representative of small sample situations. Instead we train the same settings on each fold. The best reported result in the purely supervised case is a CNN \citep{swersky2013multi,dosovitskiy2014discriminative} whose hyper parameters have been automatically tuned using 4000 images for validation achieving 70.1$\%$ accuracy. The other competitive methods on this dataset utilize the unlabeled data to learn in an unsupervised manner before applying supervised methods. We also evaluate on the full training set of 5000 images obtaining an accuracy of $87.6\%$, which is quite higher than $81.3\%$ \cite{hoffer2016deep} using unsupervised learning and the full training set. These techniques add several hyper parameters and require an additional engineering process. Applying a hybrid network is on the other hand straightforward and is very competitive with all the existing approaches without using any unsupervised learning. 
In addition to showing that hybrid networks perform well in the small sample regime, these results, along with our unsupervised CIFAR-10 result, suggest that completely unsupervised feature learning on image data may still not outperform supervised methods and pre-defined representations for downstream discriminative tasks. One possible explanation is that in the case of natural images, unsupervised learning of more complex variabilities than geometric ones (e.g the rototranslation group) might be ill-posed.
%surpasses both the supervised and unsupervised state of the art by a wide margin obtaining an average accuracy of $86.5\%$, demonstrating both the generic nature of the scattering representation and the ease of further improving the scattering representation with a hybrid architecture. %Unsupervised representation could be as well learned from a predefined scattering transform, yet this would be the topic of future research.

%% file: conclusion.tex
This work demonstrates a competitive approach for large scale visual tasks, based on scattering networks, in particular for  ILSVRC2012. When compared with unsupervised representations on CIFAR-10 or small data regimes on CIFAR-10 and STL-10, we demonstrate state-of-the-art results. We build a supervised Shared Local Encoder (SLE) that permits the scattering networks to surpass other local encoding methods on ILSVRC2012. This network of just 3 learned layers permits a deteailed analysis of the performed operations. We additionally prove that it is possible to synthetize images from a GAN in the Scattering space. % in this 3 layers network that we analyzed. 

Our work also suggests that pre-defined features are still of interest and can provide valuable insights into deep learning techniques and to allow them to be more interpretable. Combined with appropriate learning methods, they enable stronger theoretical guarantees, which are necessary to engineer better deep models and stable representations.